\documentclass[final,5p,times,twocolumn]{elsarticle}

\usepackage{amssymb}
\usepackage{amsmath} 
\usepackage{algpseudocode}
\usepackage{algorithm}

\usepackage{soul}

\usepackage{caption}
\usepackage{subcaption}
\usepackage{tabularx}
\usepackage{booktabs}
\usepackage{multirow}
\usepackage{siunitx}
\usepackage{graphicx}
\usepackage[export]{adjustbox}

\usepackage{float}

\newtheorem{definition}{Definition}

\journal{Aerospace Science and Technology}

\begin{document}

\begin{frontmatter}




\title{An Integrated Imitation and Reinforcement Learning Methodology for Robust Agile Aircraft Control with Limited Pilot Demonstration Data}


\author[inst1]{Gulay Goktas Sever\corref{cor1}\corref{cor2}} 
\ead{sever17@itu.edu.tr}
\author[inst1]{Umut Demir\corref{cor2} } \ead{demiru@itu.edu.tr}
\author[inst1]{A. Sadik Satir\corref{cor2}} \ead{sadiksatir@gmail.com}
\author[inst2]{Mustafa Cagatay Sahin}
\ead{mustafacagatay.sahin@tai.com.tr}
\author[inst1]{Nazım Kemal Ure}
\ead{ure@itu.edu.tr}

\cortext[cor1]{Corresponding author}
\cortext[cor2]{Equal contribution}

\affiliation[inst1]{addressline={Istanbul Technical University, Artificial Intelligence and Data Science Application and Research Center}, 
            city={Istanbul},
            postcode={34467}, 
            country={Turkey}}

\affiliation[inst2]{addressline={Turkish Aerospace Industries, Inc.}, 
            city={Ankara},
            postcode={06980}, 
            country={Turkey}}

\begin{abstract}

In this paper, we present a methodology for constructing data-driven maneuver generation models for agile aircraft that can generalize across a wide range of trim conditions and aircraft model parameters. Maneuver generation models play a crucial role in the testing and evaluation of aircraft prototypes, providing insights into the maneuverability and agility of the aircraft. However, constructing the models typically requires extensive amounts of real pilot data, which can be time-consuming and costly to obtain. Moreover, models built with limited data often struggle to generalize beyond the specific flight conditions covered in the original dataset. To address these challenges, we propose a hybrid architecture that leverages a simulation model, referred to as the source model. This open-source agile aircraft simulator shares similar dynamics with the target aircraft and allows us to generate unlimited data for building a proxy maneuver generation model. We then fine-tune this model to the target aircraft using a limited amount of real pilot data. Our approach combines techniques from imitation learning, transfer learning, and reinforcement learning to achieve this objective. To validate our methodology, we utilize real agile pilot data provided by Turkish Aerospace Industries (TAI). By employing the F-16 as the source model, we demonstrate that it is possible to construct a maneuver generation model that generalizes across various trim conditions and aircraft parameters without requiring any additional real pilot data. Our results showcase the effectiveness of our approach in developing robust and adaptable models for agile aircraft.

\end{abstract}




\end{frontmatter}


\section{Introduction}
"Pilot-in-the-Loop" simulation studies are a vital part of aircraft design and development process \cite{ackerman2014pilot}. High-thrust agile aircraft are evaluated according to their maneuverability and agility metrics~\cite{liefer1992fighter}, which can be obtained via flight simulators in earlier stages of aircraft design~\cite{lone2013pilot}. The high-fidelity "pilot-in-the-loop" simulators offer a controlled environment in which scenarios and algorithms can be tested without the risks associated with full-scale manned aircraft flight testing.

That being said, analyzing these capabilities requires highly skilled pilots due to the extreme flight conditions that push the maneuverability limits of the aircraft. Scheduling simulator time for such highly qualified pilots is a time and cost critical problem, since a wide range of scenarios under various trim conditions needs to be tested \cite{ackerman2014pilot,vidakovic2021flight}, and pilots with such skill set are both few in numbers and high in demand. An alternative solution is replacing the human pilot with a model built by using a combination of flight data and domain knowledge, so that the model can be used in cases where a human pilot is either not available or flying a human pilot would be high cost.

The primary objective of the maneuver generation model is to generate trajectories that closely align with the demonstrations of an expert pilot for a specific aircraft model. Additionally, the model should exhibit robustness against slight deviations from the desired course. As an inherent part of the aircraft design process, the parameters of the simulated aircraft model are subject to change. These changes encompass various aspects, including weight distribution, actuator models, low-level control parameters, and design parameters. Consequently, the model needs to possess the capability to adapt to these changes in the existing aircraft model without necessitating the acquisition of new demonstrations. Moreover, the model should exhibit transferability to similar aircraft, enabling its application across a range of aircraft models without the need for extensive re-engineering.


The main objective of this paper is to propose a methodology for designing a high-performance maneuver generation model that can effectively adapt to changes in aircraft parameters and be readily applied to new aircraft models. To achieve this goal, we leverage various tools from modern machine learning, including imitation learning, transfer learning, and reinforcement learning, in combination with existing domain knowledge on flight mechanics and aircraft dynamics.

\subsection{Previous Work}

Broadly speaking, methods for maneuver generation modeling can be categorized into two main approaches: traditional control system design-based methods and data-driven learning-based methods. In this subsection, we provide a comprehensive review of prior works in both categories.

\subsubsection{Conventional Control System Design Based Methods}

Conventional flight control system design methods are suitable for basic tasks like altitude maintenance or waypoint tracking~\cite{stevens2015aircraft}. However, controlling highly agile aircraft, especially for aerobatic and combat maneuvers, demands greater complexity and precision in control laws and aircraft models. Levin et al. \cite{levin2017agile} formulated a control policy that combines feedback and feed-forward techniques for position, attitude, speed, and altitude tracking in fixed-wing UAVs, with a focus on the knife-edge maneuver using a Rapidly-Exploring Random Tree (RRT) based planner. Further, \cite{levin2019agile} introduces a method for automating agile maneuvers in small UAVs. This method uses offline optimal control problem-solving to create reference trajectories and feedforward controls, bolstered by dynamic time warping for robustness, and feedback controls to stabilize around the reference trajectory, proving effective in various maneuvers.

Bulka and Nahon \cite{bulka2017autonomous, bulka2019automatic} introduced a physics-based control system for agile maneuvers in fixed-wing UAVs with modules for maneuver generation and both position and attitude control. McConley et al. \cite{mcconley2000hybrid} derived maneuvers from recorded pilot inputs, segmenting trajectories into predefined curves. Ure and Inalhan \cite{ure2008design, ure2012autonomous} presented a multi-modal framework that decomposes maneuvers into individual modes and parameters. This method simplifies control by targeting each sub-maneuver's feasibility and control issues separately and using specific controllers for each mode, allowing the system to track a series of maneuver modes rather than one universal controller.

\subsubsection{Learning Based Methods}
Conventional control system designs require an accurate aircraft physical model, especially for high agility maneuvers, and often don't match the agility of expert pilots. Although skilled pilots surpass conventional systems in agility, they are limited by bandwidth constraints \cite{mcconley2000hybrid}. Learning-based methods, which utilize demonstration data instead of predefined models, can emulate expert actions, enhancing performance. These techniques are especially useful where explicitly programming desired behaviors, like modeling pilot actions, is challenging. The growing body of literature on learning-based control underscores the rising interest and research in this domain, with many studies focusing on refining control of aerial vehicles \cite{ASTC1, ASTC2, ASTC3, ASTC4}.

Shukla et al. \cite{shukla2020imitation} employed the DAgger algorithm to train an ANN model to control a simulator's fixed-wing aircraft, successfully following a square flight track, though aerobatic maneuvers and model transferability weren't addressed. Medeiros \cite{medeiros2021learn} used behavior cloning to train a model on aerobatic maneuvers, specifically the Immelmann turn, using pilot demonstrations. Using a Microsoft Flight Simulator plugin for data, the LSTM model achieved the maneuver at various altitudes smoothly, without human pilot-like oscillations. Transfer learning in machine learning is well-established \cite{yosinski2014transferable, tan2018survey, zhuang2020comprehensive, iman2023review}, but transferring agile maneuvering models remains underexplored. While Medeiros's model handled aerobatics, it wasn't tested for transferability. Conversely, Sandstrom et al. \cite{sandstrom2021efficiency, sandstrom2022fighter} demonstrated model transferability across flight simulators with differing flight dynamics, focusing on human-like flight behavior. Their input vector's design required less training data and adapted more quickly to flight dynamics changes. Lastly, adapting pilot models to aircraft parameter changes is critical, as parameters vary during design. Although reinforcement learning (RL) \cite{sutton2018reinforcement} can address this without new pilot data and has been applied to uncertain aircraft models \cite{koryakovskiy2018model}, its use for learning pilot models amid parameter changes remains uncharted.

\subsection{Contributions}

Based on our discussion, the primary requirements for the maneuver generation model encompass i) robustness (consistent performance across different flight conditions), ii) transferability (ability to transfer the model to different aircraft models), and adaptation to changes in aircraft parameters (ability to adapt to changes in aircraft parameters without requiring additional data). While several studies have addressed aspects of maneuver generation modeling, there is currently no single approach that fulfills all three requirements. Notably, the robustness of the model in the presence of states absent from the demonstration data has not been extensively discussed in previous works such as \cite{medeiros2021learn, sandstrom2021efficiency, sandstrom2022fighter}, and the focus on transferability is primarily emphasized in \cite{sandstrom2021efficiency, sandstrom2022fighter} through the utilization of pre-trained models. Our primary objective in this work is to develop an approach that effectively meets all three requirements of robustness, transferability, and adaptation to changes in aircraft parameters.

\begin{figure*}[!ht]
    \centering
    \includegraphics[width=\textwidth]{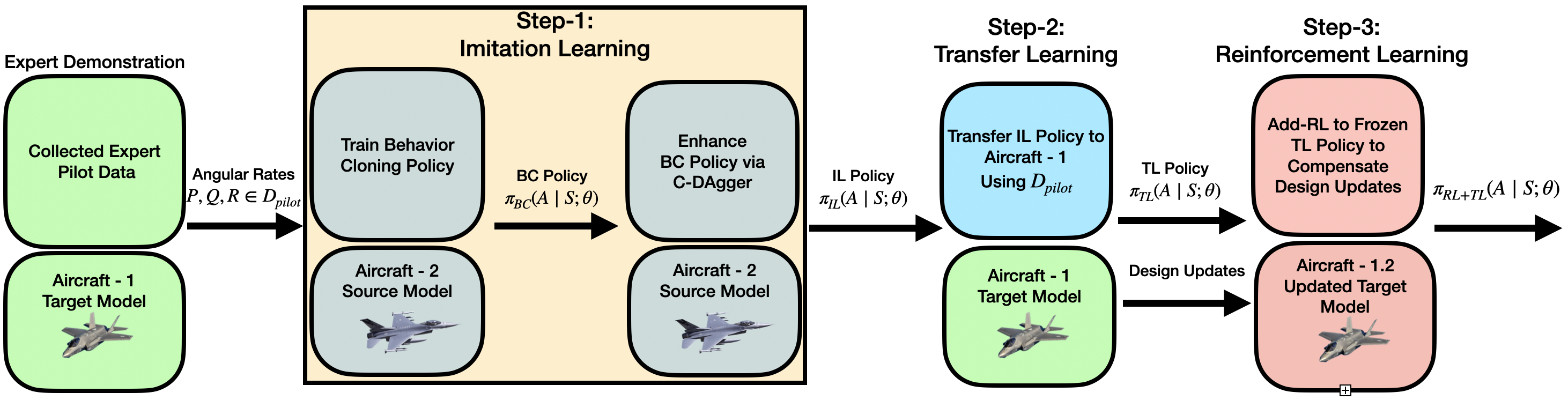}
    \caption{
   Our work outlines a three-step maneuver generation modeling process. First, we gather initial pilot demonstrations from a target aircraft model. Then, in Step 1, we apply our C-Dagger imitation learning method, which learns a behavior policy from these demonstrations using a simulated source aircraft model. In Step 2, transfer learning methods adapt this model to the target aircraft, accommodating different dynamics or characteristics. Lastly, Step 3 employs reinforcement learning techniques to further generalize the model, ensuring adaptability and robustness to variations in aircraft parameters.
    }
    \label{fig:problemdefinition}
\end{figure*}

Our proposed methodology is outlined in Fig. \ref{fig:problemdefinition}. The contributions of the paper focus on three main features:

\begin{itemize}
    \item \textbf{Robust Models:} Model robustness in the context of imitation learning refers to the agent's ability to recover and perform well in unvisited states during training. To address this, the paper introduces the Confidence-DAgger (C-Dagger) algorithm, an extension of the DAgger algorithm. DAgger involves aggregating modified training data labeled with inputs and outputs from an expert policy to mitigate cascading and compounding error problems in sequential decision-making. C-Dagger enhances the training procedure and performance of the imitation learning policy by selecting the most appropriate demonstration data. This leads to improved efficiency and robustness in the learned pilot model's performance.
    
    \item \textbf{Model Transfer to Different Aircraft:} Transfer learning plays a crucial role in addressing the issue of insufficient training data when it comes to maneuver generation modeling. By leveraging a pre-trained model, transfer learning enables the learning of a policy with limited training data. This approach is particularly advantageous in the context of maneuver generation modeling, as collecting data for different aircraft models necessitates a significant number of expert pilot demonstrations, which can be costly and time-consuming. In this paper, we demonstrate the effectiveness of transfer learning by successfully transferring our policy to another aircraft model using just a single pilot demonstration data. This efficient utilization of transfer learning showcases its potential for enabling effective pilot behavior modeling while minimizing the need for extensive data collection on different aircraft models.

    \item \textbf{Model Adaptation to Parameter Changes:}~With the active transferred imitation learning policy, the RL agent effectively learns the appropriate policy by interacting with the environment and making informed decisions in response to aircraft design changes. This framework is referred to as additive reinforcement learning method. The additive structure is particularly effective in addressing the complexity of the search space involved in aircraft motion. Simulation studies have demonstrated that a policy learned on a lighter aircraft can rapidly adapt to a heavier aircraft. Notably, the adapted policy can successfully perform the same aerobatic maneuvers on the new aircraft model without requiring any additional demonstrations. This highlights the model's ability to generalize and adapt to variations in aircraft design, showcasing its practical significance in real-world applications.
    \end{itemize}
    
To validate the effectiveness of our methodology, we make use of actual pilot data generously provided by Turkish Aerospace Industries (TAI). By utilizing the F-16 as the source model, we successfully demonstrate the feasibility of constructing a model that can effectively generalize across different trim conditions and aircraft parameters. Importantly, this achievement is accomplished without the need for any additional real pilot data. Our results serve as a compelling showcase of the efficacy of our approach in developing resilient and flexible models suitable for aircraft.

    In the remainder of the paper, we provide further details on the background information, models, maneuvers, and algorithms utilized in our study. This information is presented in Section \ref{s:back}. Subsequently, we introduce the key components of our proposed methodology in their respective subsections. Specifically, Section \ref{s:imitation} elaborates on the C-DAgger algorithm for imitation learning, Section \ref{s:transfer} demonstrates the transfer learning approach to different aircraft models, and Section \ref{s:rl} outlines the additive reinforcement learning methodology for adaptation to varying parameters.

It is important to note that we have chosen to present the numerical results of each algorithm in their corresponding sections. This decision aims to emphasize the individual contributions of each algorithm and highlight the sequential nature of the proposed methodology.

\section{Problem Definition}
\label{s:back}
\subsection{Problem Definition}

We consider the following problem in this paper: given a static data set of expert pilot demonstrations obtained from an aircraft from various different conditions, how can we build a model of the pilot that not only imitates the maneuvers demonstrated in the data set, but also generalizes across different trim conditions and parameters in the same aircraft. We will refer to this aircraft as the "target aircraft" from now on.

The main challenge in this problem is the fact that in many real-life scenarios it is infeasible to perform data-intensive operations on the target aircraft (such as using imitation learning or reinforcement learning algorithms), either because of hardware limitations and/or time constraints. We make this notion more precise with the following definition. 

\begin{definition}
\textit{Target Model}: The target model refers to the aircraft from which the demonstration data is collected and for which the pilot model is intended to be deployed with specific performance and robustness guarantees. We assume that the target model is treated as a black-box model, meaning that only the input-output relationships can be observed.
\end{definition}

Since building a robust model solely using the static demonstration data from the target model is not feasible, we introduce the concept of a source model, which is different from the target aircraft but provides unlimited simulation data.

\begin{definition}
\textit{Source Model}: The source model represents an aircraft with dynamics similar to the target model and allows for unlimited simulation data generation.
\end{definition}

Consequently, we reframe the problem as follows:

\begin{definition}
\textit{Pilot Modeling\footnote{We would like to clarify that pilot model in this context bears no relation to aspects of human factors.} Problem}: Given a static dataset of expert pilot demonstrations collected from a target model and unrestricted access to a source model, the goal is to develop a model that can imitate and generalize the maneuvers exhibited in the demonstration data across various trim conditions.
\end{definition}

To tackle this problem, we employ a combination of imitation learning (Section \ref{s:imitation}), transfer learning (Section \ref{s:transfer}), and reinforcement learning (Section \ref{s:rl}) algorithms. The integrated approach is illustrated in Figure \ref{fig:problemdefinition}.

\section{Pilot Imitation Model}
\label{s:imitation}

\subsection{Pilot Behavior Cloning}

In recent years, the aerospace industry has witnessed a surge in the adoption of deep learning techniques, including deep reinforcement learning (DRL) and imitation learning (IL) \cite{he2021explainable,wu2022learning}. Specifically, imitation learning has garnered significant attention as a means to effectively and intuitively program autonomous behavior \cite{hussein2017imitation,osa2018algorithmic}. 
Behavior cloning (BC) is a variation on IL that enables a robot or artificial intelligence (AI) system to learn a task by imitating human experts or a dataset of demonstration examples. It offers the advantage of efficient and rapid learning of new tasks without the need for extensive programming or manual intervention. BC is particularly effective in teaching tasks that require precise timing or coordination, which may be challenging for manual instruction, such as aerobatic maneuvers.

The objective of this section is to develop a pilot behavior model capable of autonomously executing agile maneuvers across the entire flight envelope, replacing the need for expert pilot demonstrations. To accomplish this objective, we introduce the concept of pilot behavior cloning and robust pilot behavior cloning through the utilization of C-DAgger, a modified version of the DAgger algorithm \cite{ross2011reduction}. C-DAgger addresses the challenges of robustness and generalization in imitation learning, enhancing the performance of the pilot behavior model.

 \subsubsection{Data Collection}
 The acquisition of diverse and representative training data plays a crucial role in ensuring the performance of the behavior cloning model. However, in the case of the target aircraft, the availability of sufficient expert pilot demonstration data for effective utilization of behavior cloning techniques is often limited (as mentioned in Section \ref{s:back}). To address this limitation, we adopt a workaround approach. We extract the body angular rates, including roll axis angular rate $P$, pitch axis angular rate $Q$, and yaw axis angular rate $R$, from the reference expert pilot maneuvers. Subsequently, we employ an open source aircraft model (referred to as the source model in Section \ref{s:back}) in conjunction with a closed-loop controller. This combination allows us to replicate the expert pilot demonstrations across a wide range of trim conditions. For this study, the expert pilot demonstrations are generously provided by the Turkish Aerospace Agency (TAI). To replicate these demonstrations, we utilize a high-fidelity nonlinear 6-DOF F-16 model \cite{nguyen1979simulator} as the source model. The Nonlinear Dynamic Inversion (NDI) controller \cite{stevens2015aircraft} is employed to govern the dynamics of the source model, ensuring accurate replication of the expert pilot maneuvers.

\begin{figure}
 \centering
 \includegraphics[width=\columnwidth]{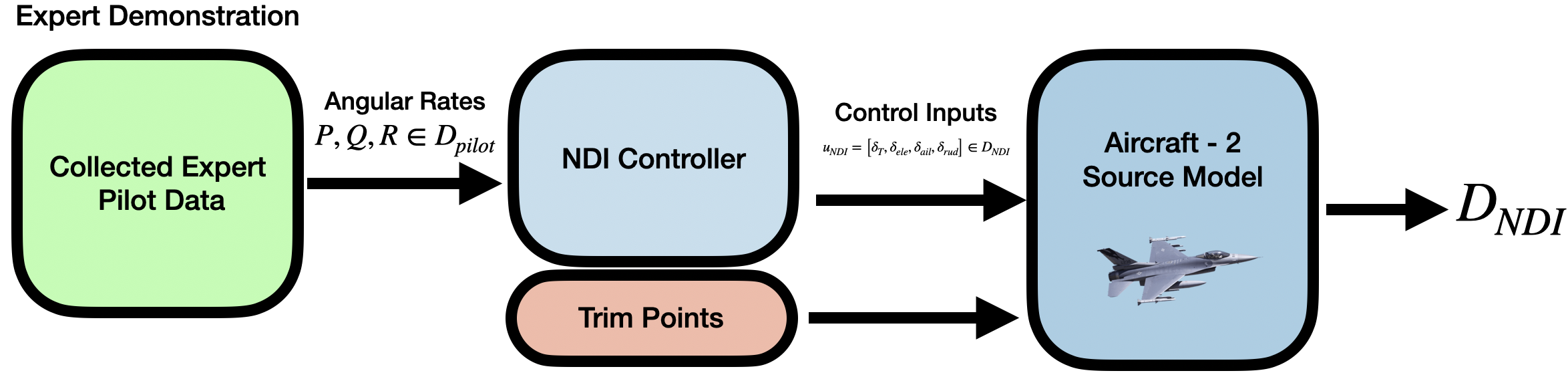} 
 \caption{Expert pilot maneuvers are employed to extract body angular rates (roll axis angular rate P, pitch axis angular rate Q, and yaw axis angular rate R) as reference signal for the Nonlinear Dynamic Inversion (NDI) controller. These maneuvers are replicated across different trim conditions using an open-source aircraft model known as the source model.}
 \label{fig:data_gen}
\end{figure}

The data generation strategy for behavior cloning is depicted in Fig. \ref{fig:data_gen}. The strategy entailed conducting simulations across a spectrum of airspeeds ($V_t$) ranging from 600 to 900 ft/s with a step size of 50 ft, as well as altitudes ($h$) spanning from 10,000 to 20,000 ft with a step size of 500 ft. In total, 56 simulations were executed, with each simulation generating 705 samples and comprising 18 states. Consequently, the cumulative number of samples amassed was 39,480.

The raw simulation data consists of 18 parameters including airspeed ($V_t$), angle of attack ($\alpha$), side-slip angle ($\beta$), roll angle ($\phi$), pitch angle ($\theta$), heading angle ($\psi$), angular rates ($P$, $Q$, $R$), and position ($p_n$, $p_e$, $p_d$), as well as control inputs ($\delta_T$, $\delta_{ele}$, $\delta_{ail}$, $\delta_{rud}$) and time ($t$). This data is represented as a state vector denoted as\\
$x_{training}=[V_t,\alpha,\beta,\phi,\theta,\psi,P,Q,R,p_n,p_e,p_d,a_n,\delta_T,\delta_{ele},\delta_{ail},\delta_{rud},t]\in\mathbb{R}^{18}$.

The  actuator limits are given as: 

\begin{align}
 \label{eq:act_lim}
 0 &\leq\delta_T\leq 1 \nonumber\\
 -25^\circ & \leq\delta_{ele}\leq 25^\circ \nonumber \\ -21.5^\circ & \leq\delta_{ail}\leq 21.5^\circ \nonumber \\
 -30^\circ & \leq\delta_{rud}\leq 30^\circ \nonumber
\end{align}, where $\delta_{T}$ is throttle position, $\delta_{ele}$ is elevator deflection, $\delta_{ail}$ is aileron deflection, $\delta_{rud}$ is the rudder deflection

\subsubsection{Neural Network Architecture}\label{sect:nn_model}

To replicate expert pilot behavior, it is crucial to employ a supervised learning algorithm capable of training an artificial neural network to directly map sensor states to control input values or actuator commands. However, the execution of agile maneuvers, such as Split-S and Chandelle, involves swift and substantial changes in control inputs, posing a challenge when designing a network that can effectively capture such dynamic behavior.

To tackle this challenge, we conducted an analysis of the control inputs involved in Split-S and Chandelle maneuvers. Specifically, we identified rapid peaks in the control inputs. During the analysis of the training set, we made an important observation regarding the data characteristics. Specifically, we noticed that the training set consisted of a high number of time series, and there was a significant correlation among them. Therefore, it became evident that the network model should possess the capability to effectively handle and accommodate these specific data characteristics.

In our work, we drew inspiration from a study conducted at Uber on time-series extreme event forecasting using neural networks \cite{laptev2017time}. This study successfully addressed the challenge of accurately predicting rare events, specifically during holidays. We found the approach presented in this study relevant to our own problem, as it shared similarities with our goal of predicting rapid changes in control commands during aerobatic maneuvers. Therefore, we adopted a composite LSTM autoencoder for sequence reconstruction and prediction, leveraging the insights and techniques from the Uber study.

\begin{figure}[H]
 \centering
 \includegraphics [width=\columnwidth,keepaspectratio]{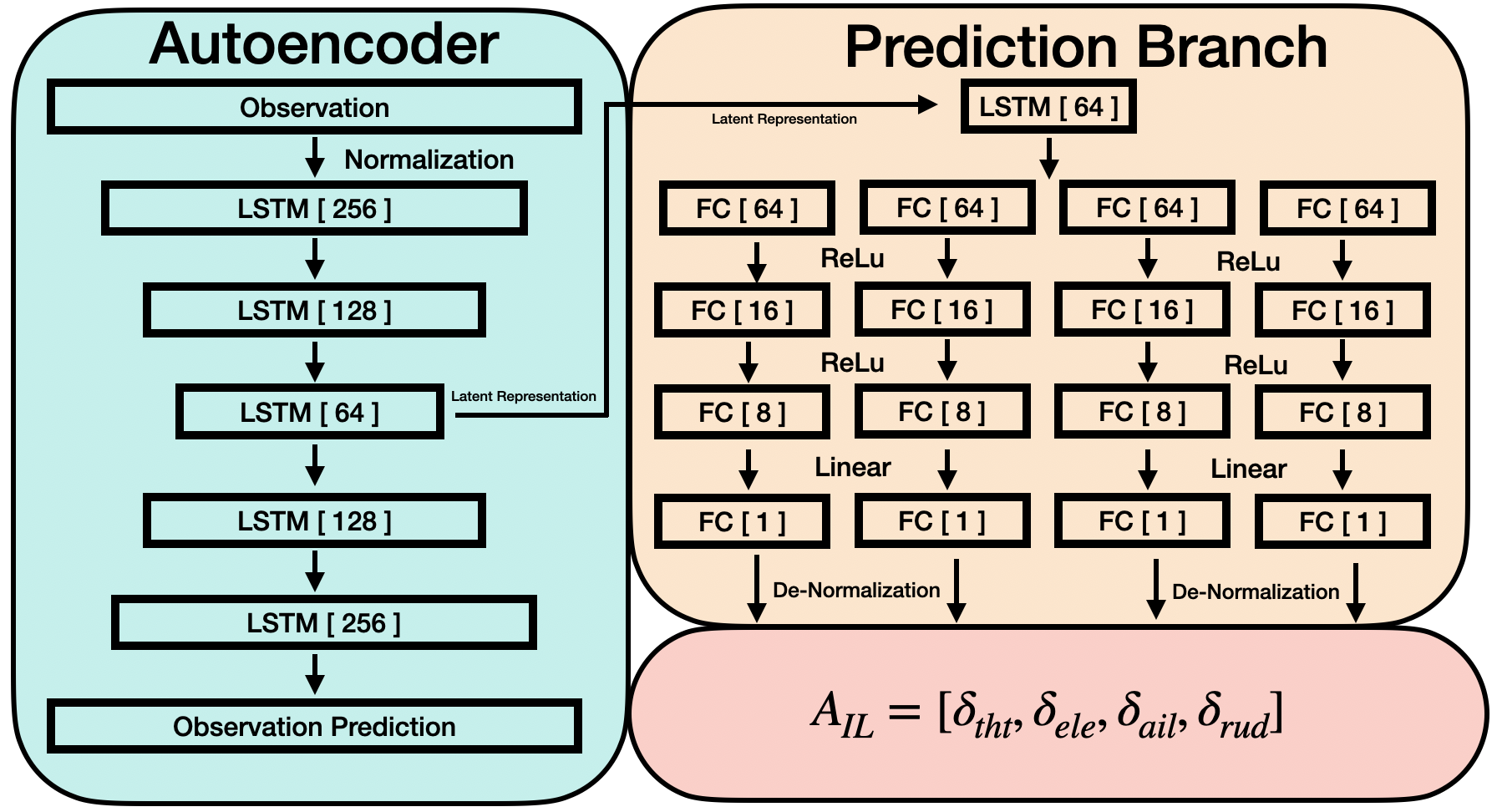} 
 \caption{Our implementation utilizes a composite LSTM autoencoder for high-dimensional aircraft data sequence reconstruction and prediction. The encoder part compresses data into a lower-dimensional latent space, capturing essential dynamics. Alongside, a prediction branch in the autoencoder leverages this latent representation to forecast future aircraft inputs.} 
 \label{fig:nn_arch}
\end{figure}

The model architecture, as depicted in Fig. \ref{fig:nn_arch}, consists of an encoder and a decoder, which together form a composite model. The purpose of this architecture is twofold: sequence reconstruction and control input prediction. The encoder takes the high-dimensional input data, representing the aircraft states, and compresses it into a lower-dimensional latent representation. This latent space captures the essential dynamics and characteristics of the aircraft's movement, enabling effective sequence reconstruction. On the other hand, the decoder reconstructs the sequence from the latent representation, aiming to reproduce the original input sequence as closely as possible. This reconstruction process helps the model learn the underlying dynamics of the data and aids in predicting control inputs accurately.

\subsubsection{Training and Evaluation}
The neural network model was trained using a supervised learning approach on the collected data. Prior to training, the data was normalized to the [0,1] interval and sampled using the sliding window method. Each input to the model consisted of a history of 50 samples for each prediction. The training data was divided into three parts: 70\% for training, 25\% for validation, and 5\% for testing.

\begin{figure}[H]
     \centering
     \begin{subfigure}[b]{\columnwidth}
         \centering
         \includegraphics[width=\textwidth]{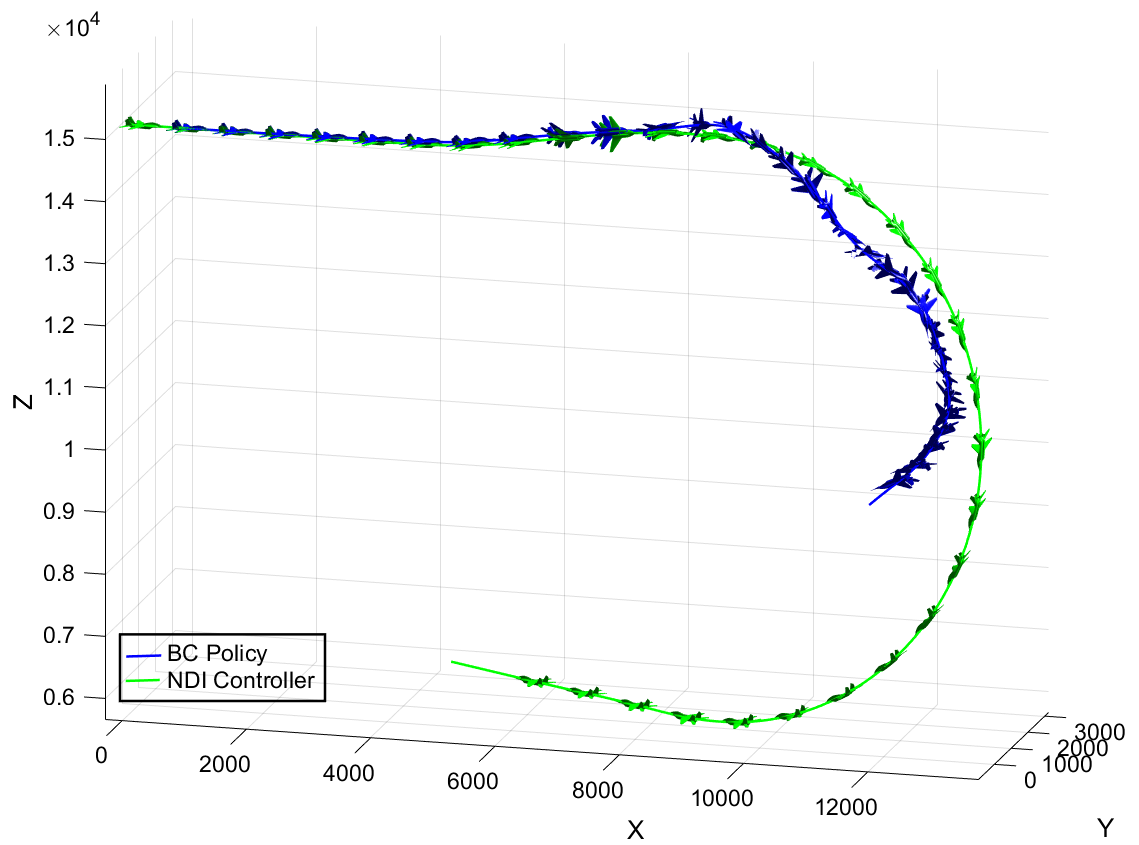}
         \caption{Split-S Maneuver}
         \label{fig:splits_bc}
     \end{subfigure}
     \hfill
     \begin{subfigure}[b]{\columnwidth}
         \centering
         \includegraphics[width=\textwidth]{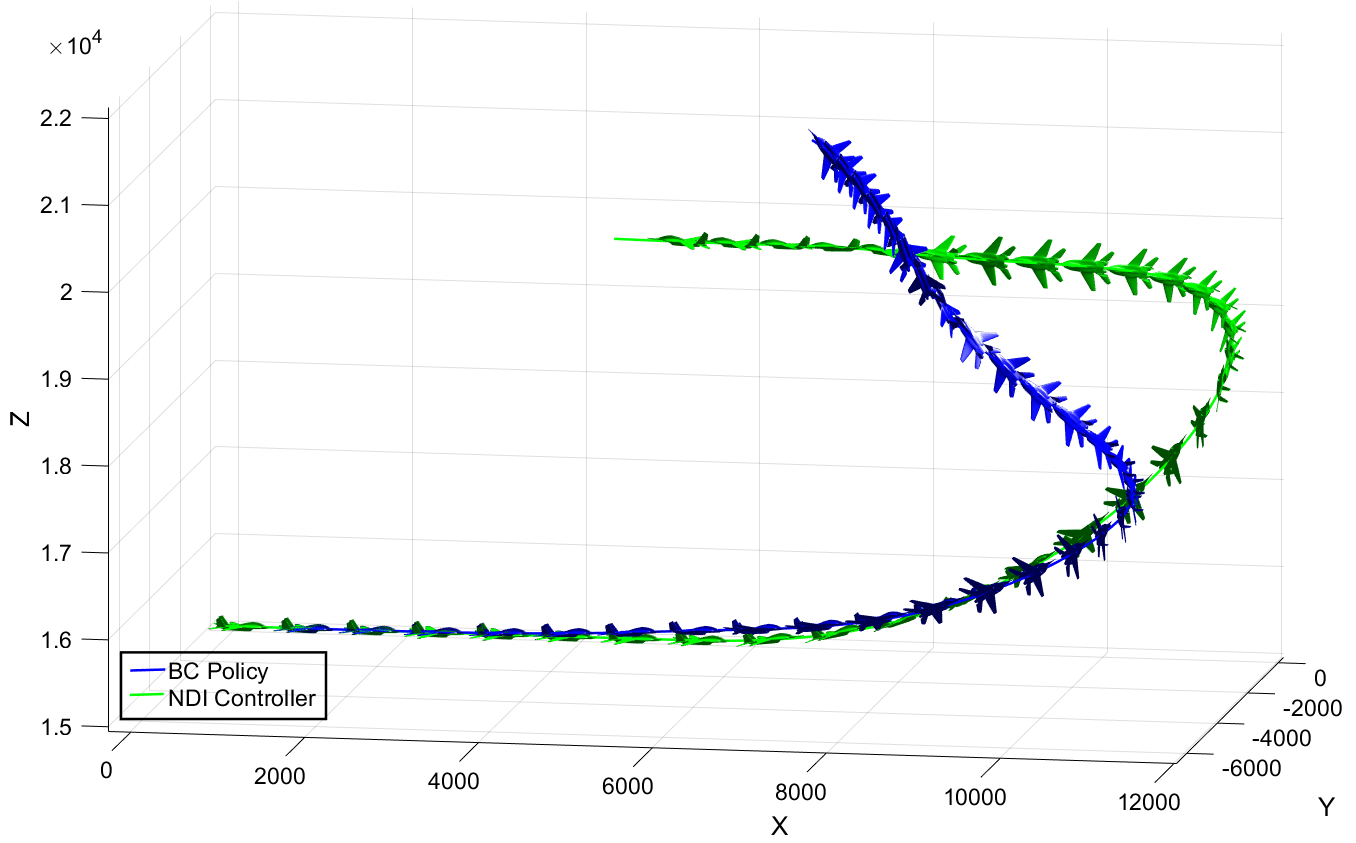}
         \caption{Chandelle Maneuver}
         \label{fig:chand_bc}
     \end{subfigure}
     \caption{To evaluate the BC policy, we executed Split-S and Chandelle maneuvers on Aircraft - 2 Source Model using BC and compared them with the NDI controller-executed trajectories. This comparison helps determine the BC policy's effectiveness in replicating desired maneuvers.}
     \label{fig:bc_manv}
\end{figure}

Behavior cloning methods utilize a loss function to quantify the disparity between the expert's behavior and the learned policy. The Mean Squared Error (MSE) loss function is commonly employed to ensure the learned policy closely replicates the expert's behavior. The underlying assumption is that a deep control policy that closely resembles a real pilot's maneuver will imitate their behavior in operational settings. In our study, BC policies for Split-S and Chandelle maneuvers were trained, and the corresponding trajectories are depicted in Fig. \ref{fig:splits_bc} and Fig. \ref{fig:chand_bc}.

It should be noted that behavior cloning (BC) policies have limitations when it comes to recovering from initial deviations and handling unvisited states that are not included in the training data. The errors made during training can compound, leading to suboptimal performance, as shown in Figures \ref{fig:splits_bc} and \ref{fig:chand_bc}. This is due to the mismatch between the training and test data sets, violating the assumption of independent and identically distributed data in supervised learning\cite{bagnell2015invitation,osa2018algorithmic}. To address these limitations, we propose a new imitation learning algorithm called C-DAgger. C-DAgger allows the model to interact with the environment and learn from its own mistakes, resulting in a more robust and generalizable pilot behavior model. In the following section, we will provide a detailed explanation of the C-DAgger algorithm and its benefits in overcoming the limitations of behavior cloning.

\subsection{Robust Pilot Behavior Cloning via Confidence DAgger}

Behavior cloning with interactive demonstrations has the advantage of allowing the model to learn from its own mistakes, improving its adaptability and generalization to new environments. To address the compounding error problem in behavior cloning, the DAgger (Data Aggregation) algorithm was proposed by Ross et al. \cite{ross2011reduction}. DAgger reduces imitation learning to supervised learning with interaction by iteratively collecting data from the agent's own actions and combining it with expert demonstrations to train a new policy. However, DAgger and similar approaches have limitations in learning from complex and high-dimensional state spaces, such as those encountered in aerobatic maneuvers. Collecting training samples from both the reference and trained policies can be costly. To mitigate this issue, sample-efficient DAgger algorithms have been proposed in the literature, such as Query-Efficient Imitation Learning \cite{zhang2017query} and Sample Efficient Interactive End-to-End Deep Learning \cite{bicer2019sample}. These algorithms aim to collect critical demonstration data while minimizing the number of queries to the reference policy, making the training process more efficient.

To address the challenges posed by the large dimension of our state-action space, it is crucial to develop a sample-efficient DAgger algorithm that intelligently selects critical demonstration data for training while avoiding the use of incorrect actions. An effective switching/selective logic between the expert policy and the trained policy plays a vital role in achieving this objective. Existing research works such as "Interactive Learning from Human-Guidance" \cite{chernova2009interactive} and "The Hindsight Experience Replay" \cite{kelly2019hg} have proposed methods for selecting appropriate demonstration data from human experts, which can be adapted to our specific problem. Inspired by these concepts, we have devised the Confidence-DAgger (C-DAgger) algorithm to enhance the robustness and generalization of our behavior cloning (BC) model in a more efficient manner. C-DAgger incorporates a confidence mechanism similar to the approaches utilized in "Interactive Learning from Human-Guidance" and "The Hindsight Experience Replay". In C-DAgger, our expert policy NDI focuses on precisely tracking the P-Q-R values that characterize successful aerobatic maneuvers. The expert policy query is based on the confidence in the $P-Q-R$ tracking errors, allowing us to effectively identify critical demonstrations for training.



The $P-Q-R$ tracking errors, which quantify the deviation in rad/s between the actual and desired PQR values, are computed as follows:

\begin{align}
\Delta_P = |P_d - P| ,\quad \Delta_Q = |Q_d - Q| ,\quad\Delta_R = |R_d - R|
\end{align}

The switching parameter $\lambda$ between the learner policy and expert policy depends on the confidence gain $c_g$.

$\lambda = \begin{cases}
            1 & \text{if } \Delta_P < P_{d}c_g \quad \text{and} \quad \Delta_Q < Q_{d}c_g \quad \text{and} \quad \Delta_R < R_{d}c_g \\
            0 & \text{otherwise}
        \end{cases}$
\\To determine the optimal confidence gain $c_g$, we maximize the cumulative summation of $\Lambda$:
\begin{align*}
    c_g\gets\underset{c_g}{\max}&\quad\Lambda=\sum_{i=1}^N\sum_{t=1}^\tau\lambda\\
    \text{s.t.}&\quad x(t+1) = \textbf{6-DOF}(x(t),u(t)) \text{  ,}
    &\quad x(t+1) \in X_{\text{stable}}
\end{align*}

The confidence gain in the C-DAgger algorithm triggers the expert policy to provide corrective demonstrations when it reaches a critical threshold. This threshold corresponds to the point where the expert policy demonstrates the highest capability in recovering from deviations from the desired trajectory across the entire flight envelope. This shared decision-making mechanism ensures that the most critical demonstration data is selected for training, while avoiding incorrect actions. As a result, C-DAgger effectively combines the benefits of confidence-based selection and the DAgger algorithm. During the training process, the algorithm continuously updates the dataset and trains the learner policy. Training iterations are performed until the average error in the roll, pitch, and yaw rates falls below a specified threshold $\epsilon_{PQR}$ across the entire flight envelope. This criterion ensures that the learner policy achieves a desired level of accuracy and performance in maneuver execution. The final trained policy $\pi^{*}$ represents a refined policy that surpasses the initial behavior cloning policy in terms of maneuver performance, demonstrating the effectiveness of the C-DAgger algorithm.




\begin{algorithm}[H]
\tiny
\caption{Confidence-DAgger (C-DAgger) algorithm}
\begin{algorithmic}[1]
\State \textbf{Input:} $\mathcal{D}_{BC}$, $\pi_{BC}$, $\pi_{NDI}$, $P_d$, $Q_d$, $R_d$, $\in \mathcal{D}_{pilot}$, $\epsilon_{PQR}$  \textbf{ Output:} Learner policy $\pi^{*}$

\State \textbf{Initialize:} Set $\mathcal{D} \gets \mathcal{D}_{BC}$, $\pi^{*} \gets \pi_{BC}$

\While{True}
\State Compute $c_g$ \Comment{The confidence gain}
\For{$i$ \textbf{from} 1 \textbf{to} $N$}
    \For{$t$ \textbf{from} 1 \textbf{to} $\tau$}
        \State Compute $\lambda$ \Comment{The switching parameter}
        \State $u(t)=\lambda\pi^*(x(t))+(1-\lambda)\pi_{NDI}(x(t))$
        \State $D=D\cup\{x(t),u(t)\}$
        \State x(t+1) = \textbf{6-DOF}(x(t),u(t))
    \EndFor
\EndFor
\State $e_{pqr} = \frac{1}{N}\frac{1}{\tau}\sum_{i=1}^N\sum_{t=1}^\tau \frac{1}{3}(\Delta_P+\Delta_Q+\Delta_R )$
\If{$e_{pqr} < \epsilon_{PQR}$}
    \State break
\Else
    \State Train the policy $\pi^{*}$ on $\mathcal{D}$.
\EndIf
\EndWhile
\end{algorithmic}
\end{algorithm}

\subsubsection{Training and Evaluation}
\begin{figure*}
     \centering
     \begin{adjustbox}{minipage=\linewidth}
     \begin{subfigure}[b]{0.49\textwidth}
         \centering
         \includegraphics[width=\textwidth]{fig/SplitS/bc/SplitS_bc.png}
         \caption{BC Policy}
         \label{fig:splits_bc_manv}
     \end{subfigure}
     \hfill
    \begin{subfigure}[b]{0.49\textwidth}
         \centering
         \includegraphics[width=\textwidth]{fig/Chandelle/bc/Chandelle_bc.png}
         \caption{BC Policy}
         \label{fig:chand_bc_manv}
     \end{subfigure}     
     \medskip
     
     \begin{subfigure}[b]{0.49\textwidth}
         \centering
         \includegraphics[width=\textwidth]{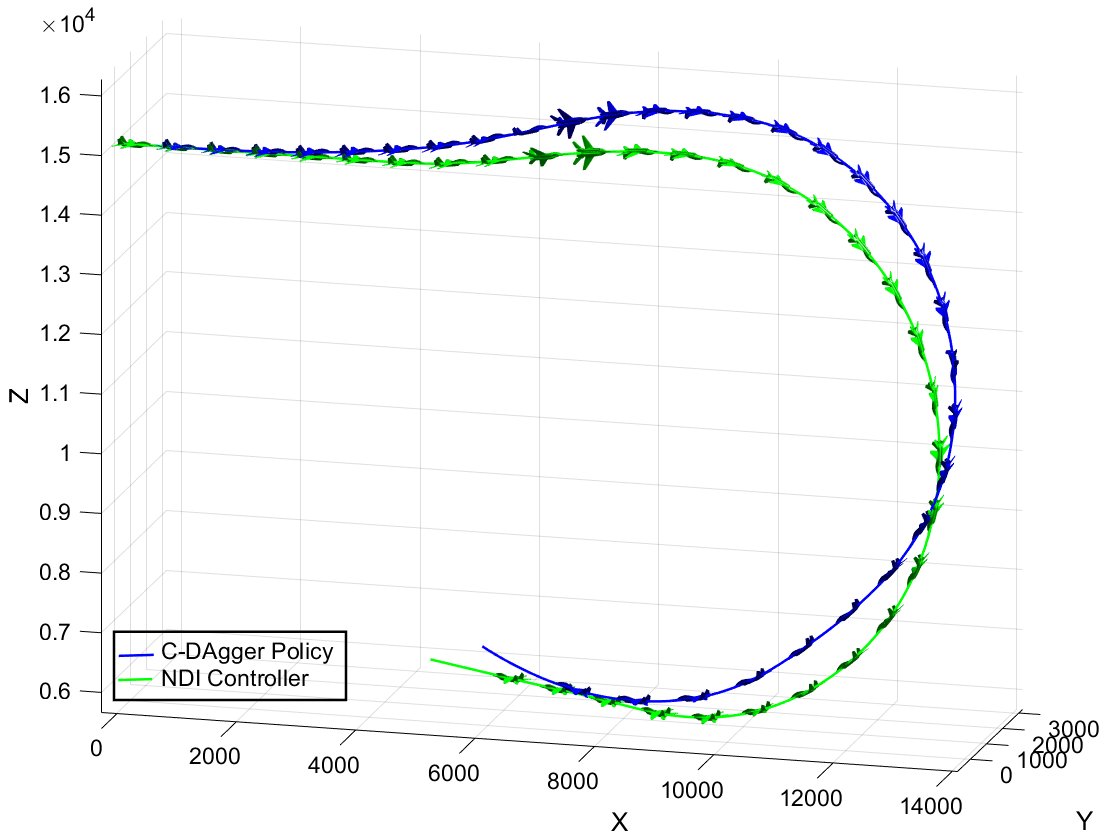}
         \caption{C-DAgger Iteration-1 Policy}
         \label{fig:splits_iter1_manv}
     \end{subfigure}
     \hfill
      \begin{subfigure}[b]{00.49\textwidth}
         \centering
         \includegraphics[width=\textwidth]{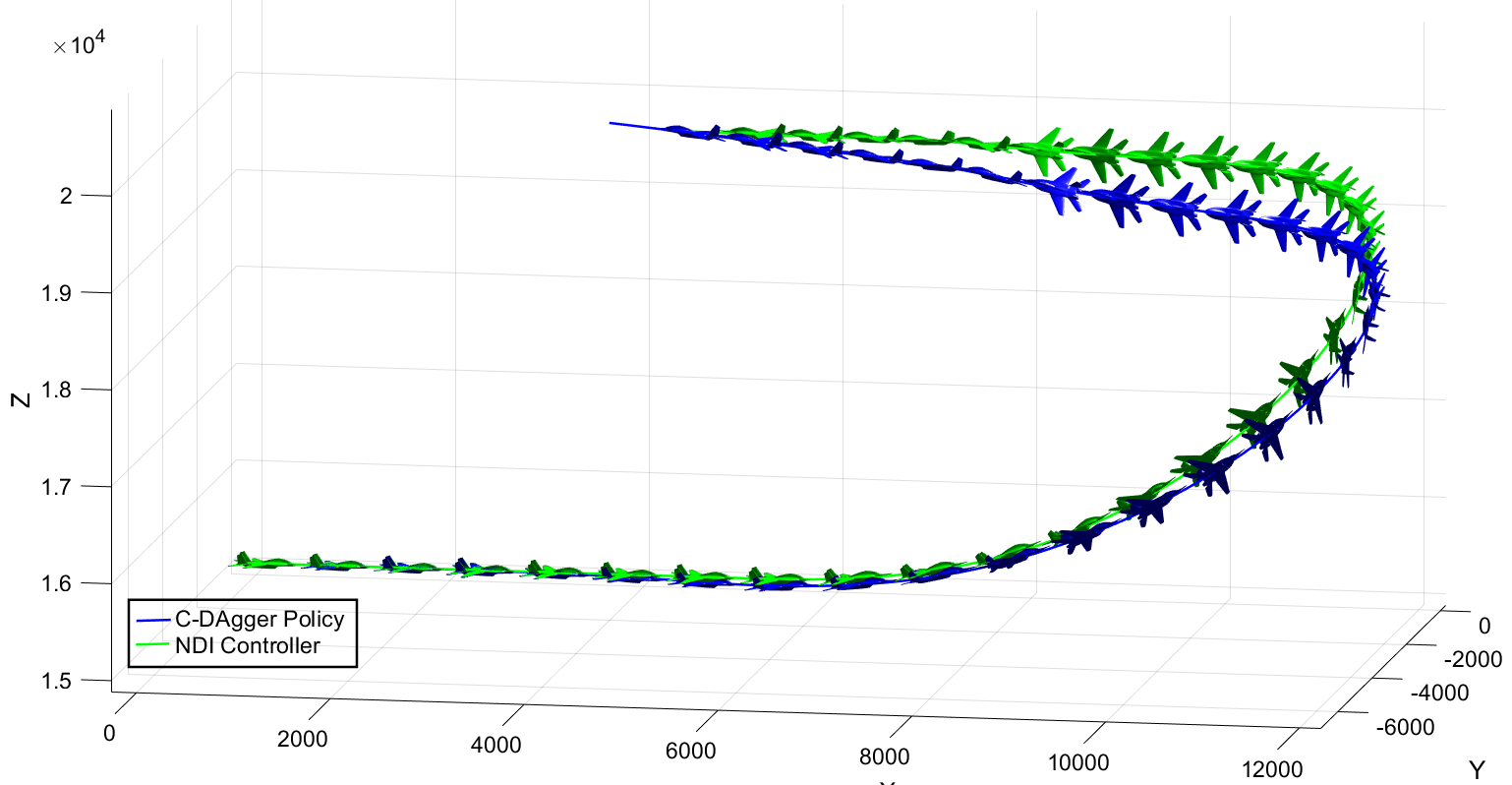}
         \caption{C-DAgger Iteration-1 Policy}
         \label{fig:chand_iter1_manv}
     \end{subfigure}
     \medskip
        \begin{subfigure}[b]{0.49\textwidth}
         \centering
         \includegraphics[width=\textwidth]{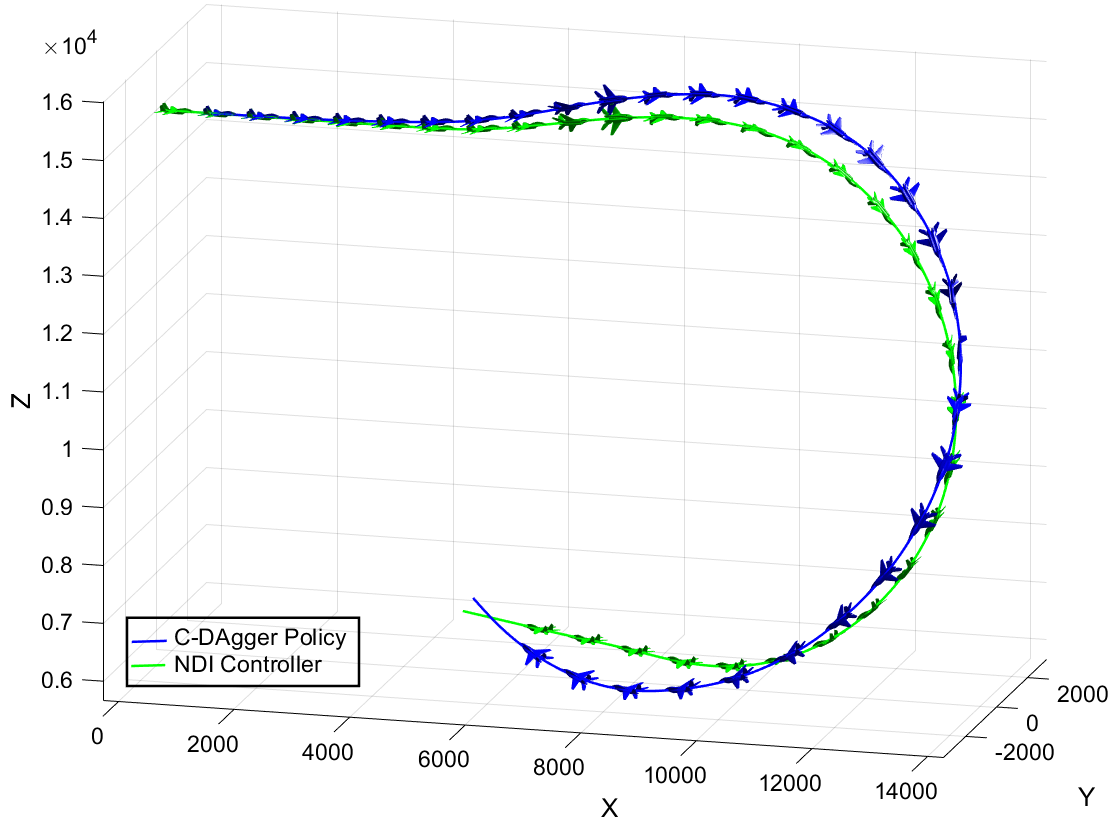}
         \caption{C-DAgger Iteration-2 Policy}
         \label{fig:splits_iter2_manv}
     \end{subfigure}
     \hfill
     \begin{subfigure}[b]{0.49\textwidth}
         \centering
         \includegraphics[width=\textwidth]{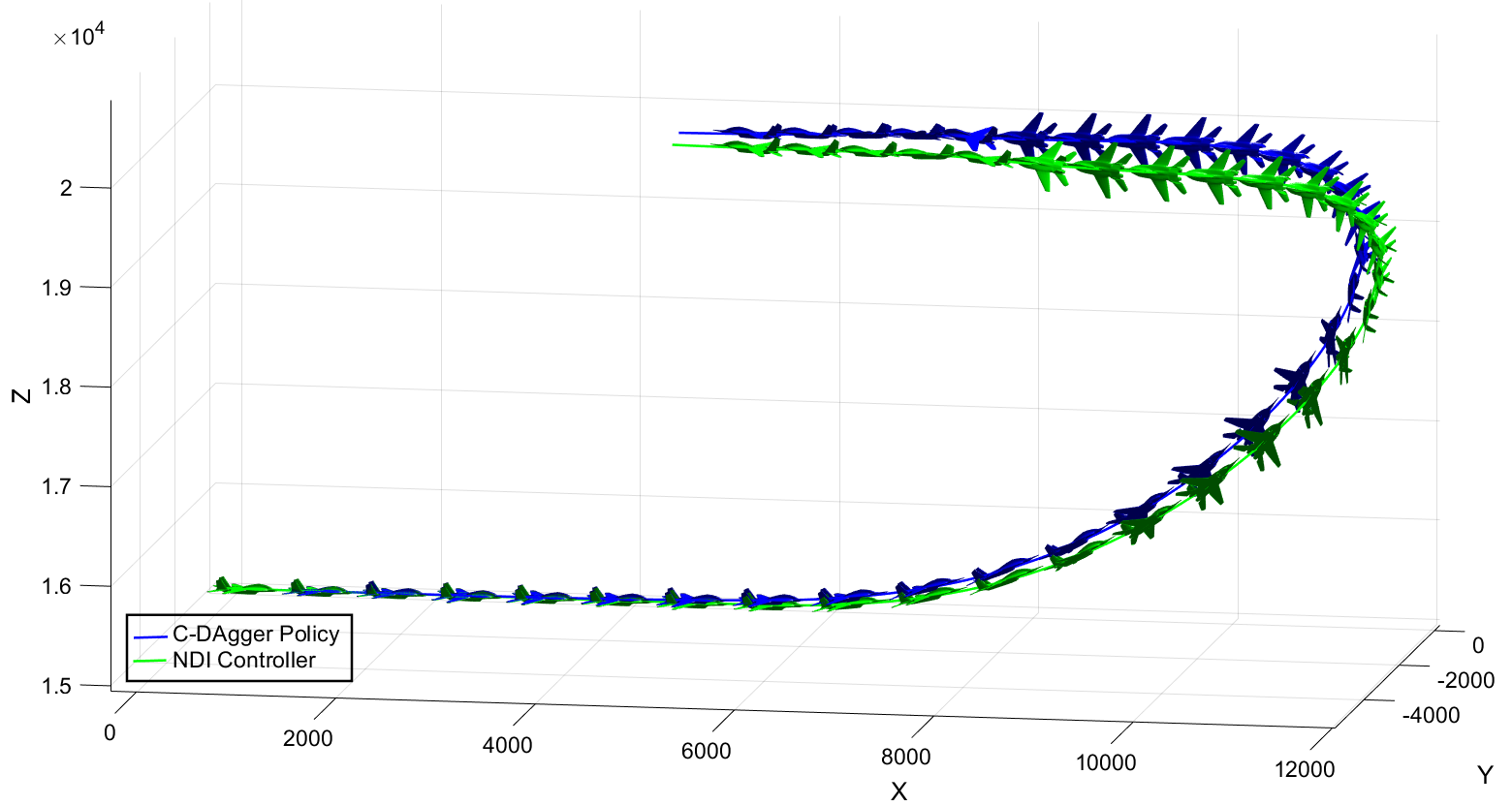}
         \caption{C-DAgger Iteration-2 Policy}
         \label{fig:chand_iter2_manv}
     \end{subfigure} 
     \end{adjustbox}
\caption{Figure shows the results of Split-S and Chandelle maneuvers in the first and second columns, respectively. Each column contains three subfigures, comparing the performance of the BC policy with NDI, and showing the results of the first and second C-DAgger iterations. The C-DAgger algorithm improves maneuver performance, as demonstrated in each row.}
 \label{fig:dagger_manv}
\end{figure*}

Fig. \ref{fig:dagger_manv} presents the results of the Split-S and Chandelle maneuvers in the first and second columns, respectively, using a developed pilot behavior model under trim conditions of velocity = 750 ft/s and altitude = 15000 ft. 

\begin{figure}[H]
     \centering
     \begin{subfigure}[b]{\columnwidth}
         \centering
         \includegraphics[width=\textwidth]{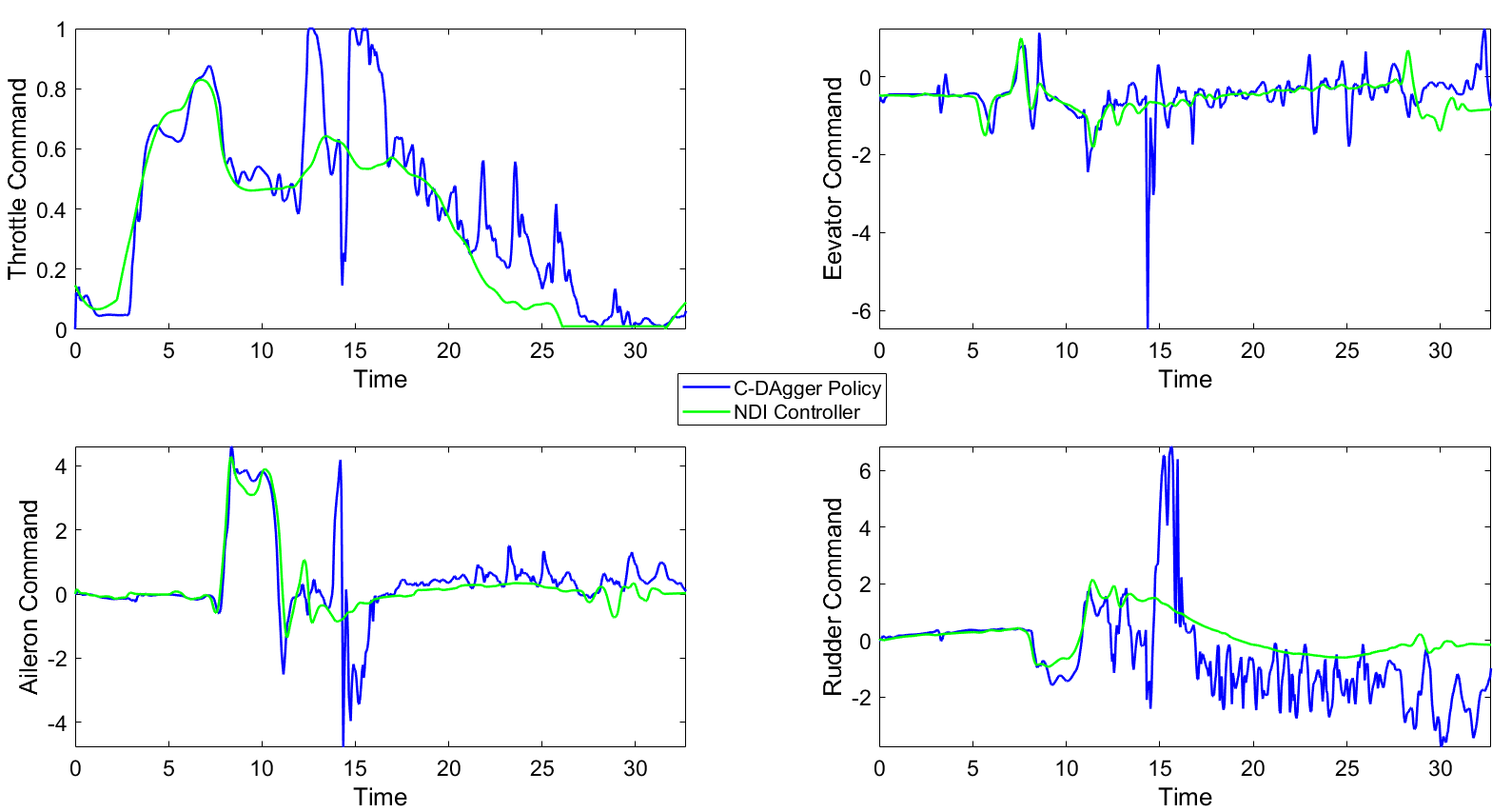}
         \caption{Split-S C-DAgger Iteration-2 Policy}
         \label{fig:splits_iter2_act}
     \end{subfigure}
     \hfill
     \begin{subfigure}[b]{\columnwidth}
         \centering
         \includegraphics[width=\textwidth]{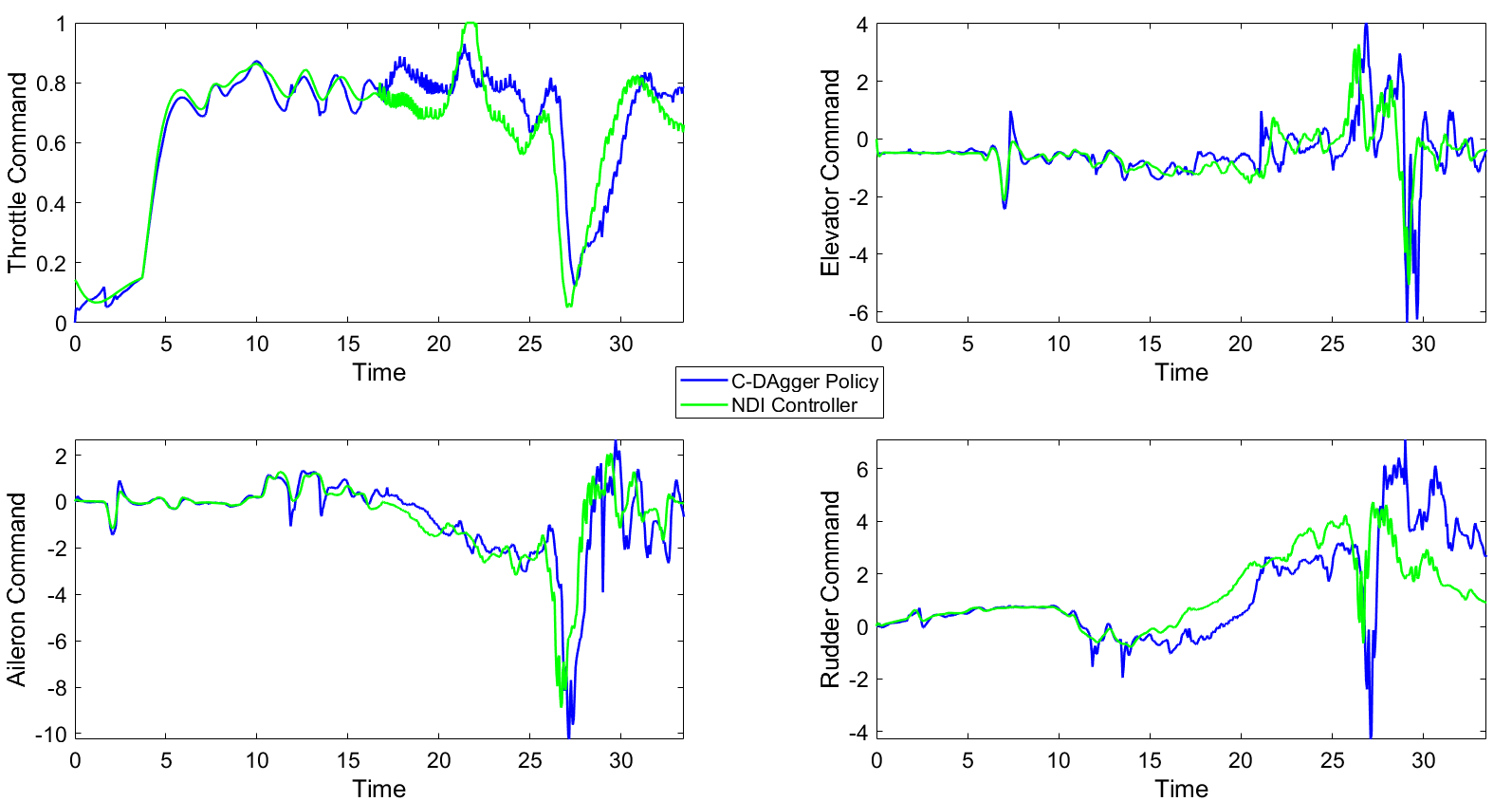}
         \caption{Chandelle C-DAgger Iteration-2 Policy}
         \label{fig:chand_iter2_act}
     \end{subfigure}
    \caption{Figure shows the normalized actuator values generated by the final learner policy obtained from the C-DAgger Iteration-2 for Split-S and Chandelle maneuvers. These actuator values conform to the specified rate and magnitude limits, demonstrating the policy's capability to generate control commands that meet the actuator limits specified in Equation-\ref{eq:act_lim}.}
     \label{fig:bc_act}
\end{figure}

The Fig. \ref{fig:bc_act} provides the actuator values for the final learner policy obtained from the C-DAgger iterations. These actuator values, namely throttle, elevator, rudder, and aileron, adhere to the rate and magnitude limits magnitude limits specified in the Eq. \ref{eq:act_lim}.
The presented actuator values are a crucial aspect of the developed model, as they reflect the policy's ability to generate control commands that meet the specified actuator dynamics requirements. 

\begin{table}[H]
\centering
\resizebox{\columnwidth}{!}{%
\begin{tabular}{|l|ccc|ccc|}
\hline
\multicolumn{1}{|c|}{} & \multicolumn{3}{c|}{\textbf{SplitS Maneuver}} & \multicolumn{3}{c|}{\textbf{Chandelle Maneuver}} \\ \hline
\multicolumn{1}{|c|}{} & \multicolumn{1}{c|}{\textbf{BC}} & \multicolumn{1}{c|}{\textbf{C-DAgger Iter-1}} & \textbf{C-DAgger Iter-2} & \multicolumn{1}{c|}{\textbf{BC}} & \multicolumn{1}{c|}{\textbf{C-DAgger Iter-1}} & \textbf{C-DAgger Iter-2} \\ \hline
\textbf{P-mse ($rad^2/s^2$)} & \multicolumn{1}{c|}{0.2402} & \multicolumn{1}{c|}{0.0334} & 0.0252 & \multicolumn{1}{c|}{0.04008} & \multicolumn{1}{c|}{0.05479} & 0.01364 \\ \hline
\textbf{Q-mse ($rad^2/s^2$)} & \multicolumn{1}{c|}{0.1555} & \multicolumn{1}{c|}{0.0119} & 0.0072 & \multicolumn{1}{c|}{0.00193} & \multicolumn{1}{c|}{0.00336} & 0.00083 \\ \hline
\textbf{R-mse ($rad^2/s^2$)} & \multicolumn{1}{c|}{0.0663} & \multicolumn{1}{c|}{0.00041} & 0.000076 & \multicolumn{1}{c|}{0.00367} & \multicolumn{1}{c|}{0.00088} & 0.00027 \\ \hline
\textbf{PQR-mse ($rad^2/s^2$)} & \multicolumn{1}{c|}{0.1540} & \multicolumn{1}{c|}{0.0152} & 0.0111 & \multicolumn{1}{c|}{0.015} & \multicolumn{1}{c|}{0.019679} & 0.00491 \\ \hline
\end{tabular}%
}
\caption{The Mean Squared Error (MSE) values for $P-Q-R$ tracking performance of the Split-S maneuver and the Chandelle maneuver for each policy are presented in this table. The learned policies compared are the Behavior Cloning (BC) algorithm, C-DAgger Iteration-1, and C-DAgger Iteration-2.}
\label{tab:pqr_track}
\end{table}

Table \ref{tab:pqr_track} presents the MSE values for the Split-S and  Chandelle  maneuver. It can be observed that with each iteration of the C-DAgger algorithm, there is a significant reduction in MSE values for all components, indicating an improvement in the accuracy of $P-Q-R$ tracking.

\section{Transfer Learning to Different Aircraft Models}
\label{s:transfer}
In the previous section we were able to obtain a robust pilot model by replicating/extending the target aircraft pilot data on the source aircraft model. Although, the resulting pilot model achieved good performance, it would only work on the source model. As it is indicated in Section \ref{s:back}, the ultimate objective is to obtain a pilot model that works with the target aircraft. For that purpose, in this section we employ transfer learning techniques to fine tune the pilot model obtained from the source aircraft model to the target aircraft model.

Transfer learning in pilot behavior modeling involves leveraging a pre-trained model obtained from a source aircraft model and adapting it to a target aircraft model using a minimal amount of additional training data. This approach utilizes model-based methods, where pre-trained layers in the network model are employed. Techniques such as freezing, fine-tuning, and adding new layers are utilized to enhance the performance of the transferred policy.

\subsection{Training and Evaluation}

\begin{figure}[H]
 \centering
 \includegraphics [width=\columnwidth]{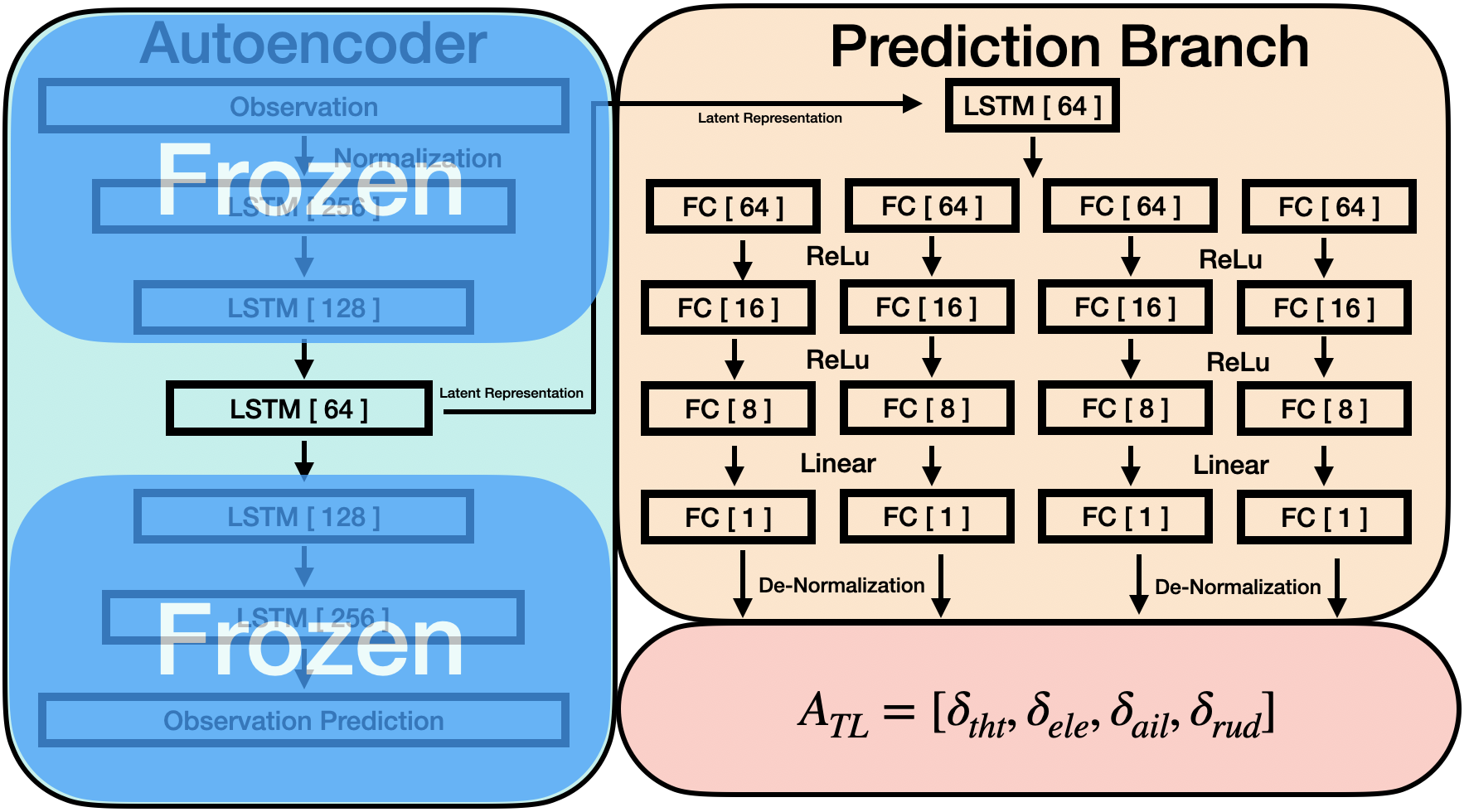} 
 \caption{ Schematic representation of the network architecture implementing freezing and fine-tuning techniques during the training process. The LSTM layers of the autoencoder, highlighted in blue, are frozen to preserve the BC model's understanding of general aircraft dynamics. The remaining layers are fine-tuned using a single demonstration data from the target domain, enabling the network to adjust and specialize for the specific aircraft.}
 \label{fig:nn_arch_tl}
\end{figure}

In our problem, we utilized freezing and fine-tuning techniques to capitalize on the advantages of a pre-trained model when training on the target data. This approach involved selectively sharing and freezing parameters in specific layers, as illustrated in Fig.  \ref{fig:nn_arch_tl} (highlighted in blue), to preserve the acquired knowledge.


The remaining layers of the network were fine-tuned using a single demonstration from the target domain, enabling the network to adapt and specialize for the specific task. The resulting policy, obtained by freezing and fine-tuning on the target data, will be referred to as the Transfer Learning (TL) policy. The training parameters for the TL policy are provided in Table \ref{tab:TLmodelparams}.


\begin{table}[H]
\centering
\resizebox{\columnwidth}{!}{%
\begin{tabular}{|l|c|l|c|}
\hline
Total number of parameters         & 1,080,214 & Learning Rate & 0.00005 \\ \hline
Trainable parameters     & 103,812   & Batch Size    & 64   \\ \hline
Non-trainable parameters & 976,402   & Epoch         & 2000 \\ \hline
\end{tabular} }
\caption{Transfer Learning Parameters}
\label{tab:TLmodelparams}
\end{table}

Despite the significant differences in flight dynamics, actuator dynamics, and design between the source model (the open source F-16) and the target model (an aircraft designed by TAI), our transfer learning (TL) policy showed remarkable success in executing the maneuver. Quantifying the disparity between these two aircraft models is a challenging task. Notably, our behavior cloning (BC) policy was trained on a dataset consisting of 39,480 samples, while the TL policy, leveraging only a single demonstration, was trained with a substantially smaller dataset of 669 samples.

\begin{figure}[H]
     \centering
     \begin{adjustbox}{minipage=\linewidth}
    \begin{subfigure}[b]{0.49\textwidth}
         \centering
         \includegraphics[width=\textwidth]{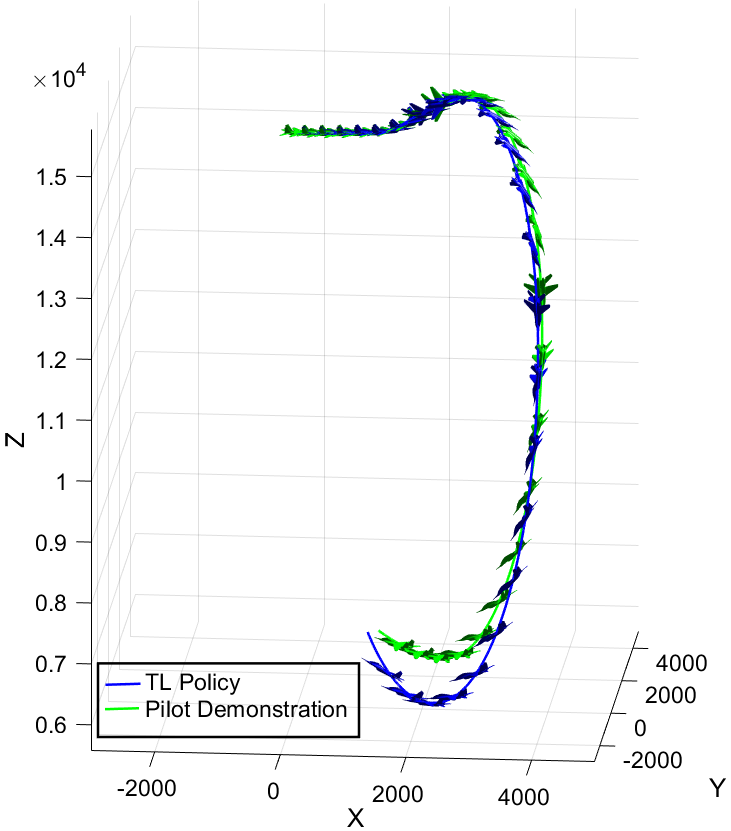}
         \caption{TL for Split-S maneuver.}
         \label{fig:splits_tl_manv}
     \end{subfigure}
     \hfill
    \begin{subfigure}[b]{0.49\textwidth}
         \centering
         \includegraphics[width=\textwidth]{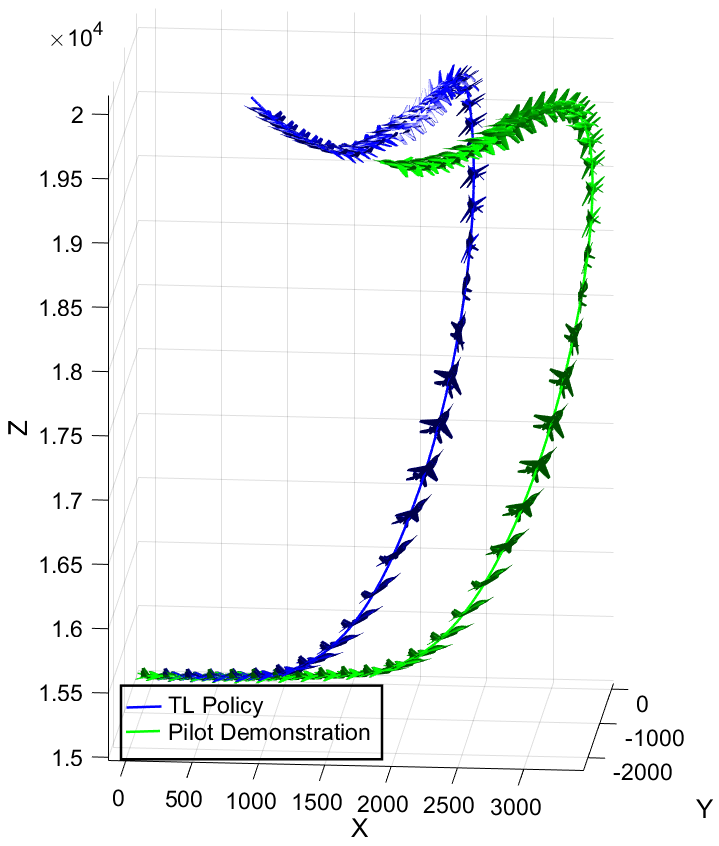}
         \caption{TL for Chandelle maneuver.}
         \label{fig:chand_tl_manv}
     \end{subfigure} 
     \end{adjustbox}
  \caption{Results of Split-S and Chandelle maneuvers on Aircraft - 1 Target Model obtained by Transfer learning (TL) policy under trim conditions of velocity = 750 ft/s and altitude = 15000 ft.} 
 \label{fig:tl_manv}
 
\end{figure}

Figure \ref{fig:tl_manv} showcases the outcomes of the Split-S and Chandelle maneuvers in the first and second columns, respectively. The simulations were conducted under trim conditions with a velocity of 750 ft/s and an altitude of 15000 ft,  which aligns with the trim point of the single demonstration data.

These results are highly promising, demonstrating the exceptional generalization performance of our source model despite the complex challenges encountered. The TL policy outperformed the BC policy in the Split-S maneuver, surpassing its performance in maneuver execution based on the $P-Q-R$ tracking metric. While the performance of the Chandelle maneuver closely matched that of the BC policy, the TL policy exhibited superior capabilities. Detailed results of the $P-Q-R$ tracking for the TL policy are presented in Table \ref{tab:tl_pqr}.

\begin{table}
\centering
\resizebox{\columnwidth}{!}{%
\begin{tabular}{|c|c|c|c|c|}
\hline
\textbf{}                        & \textbf{P-mse ($rad^2/s^2$)} & \textbf{Q-mse ($rad^2/s^2$)} & \textbf{R-mse ($rad^2/s^2$)} & \textbf{PQR-mse ($rad^2/s^2$)} \\ \hline
\textbf{Split-S Maneuver}       & 0.0125                        & 0.0030                        & 7.5693e-5                    & 0.0052                          \\ \hline
\textbf{Chandelle Maneuver}     & 0.0481                        & 0.0174                        & 0.0021                        & 0.0226                          \\ \hline
\end{tabular}
}
\caption{
Table displays the $P-Q-R$ tracking results for the Transfer Learning (TL) policy. It demonstrates exceptional generalization performance, surpassing the Behavior Cloning (BC) policy in the Split-S maneuver while closely matching its performance in the Chandelle maneuver.}
\label{tab:tl_pqr}
\end{table}

In conclusion, our study successfully transfers the imitation learning (IL) model to the target using just one demonstration, addressing the challenge of frequent pilot data collection in aircraft design. While effective, our system's adaptability can be improved. To advance this, we aim to build upon the transfer learning (TL) policy and incorporate reinforcement learning techniques, combining the strengths of both IL and RL for enhanced pilot behavior modeling in agile aircraft.

\section{Reinforcement Learning for Adaptation to Model Updates}
\label{s:rl}

In previous sections, we developed a pilot behavior model on source aircraft model using imitation learning. We then applied transfer learning to transfer the model to target aircraft model using pilot demonstration data. However, when the model is intended to be used on a target aircraft without pilot data or knowledge of the aircraft's dynamics, its performance may be compromised due to the differences in the dynamics between the learned and updated aircraft model. These small variations in dynamics can lead to significant behavioral changes in the model's performance. To address this limitation, we propose a reinforcement learning (RL) method that can be utilized in conjunction with the transfer learning (TL) policy. By integrating TL policy into the reinforcement learning framework, we aim to capitalize on the acquired knowledge from the TL policy and enable our system to continuously improve and adapt to evolving aircraft dynamics, eliminating the need for extensive data collection. This approach holds tremendous potential in achieving robust and efficient pilot behavior modeling in the context of our problem.

\subsection{The Transfer Learning Model on Updated Target Aircraft}

\begin{figure}
     \centering
     \begin{adjustbox}{minipage=\linewidth,scale=0.95}
     \begin{subfigure}[b]{0.49\textwidth}
         \centering
         \includegraphics[width=\textwidth]{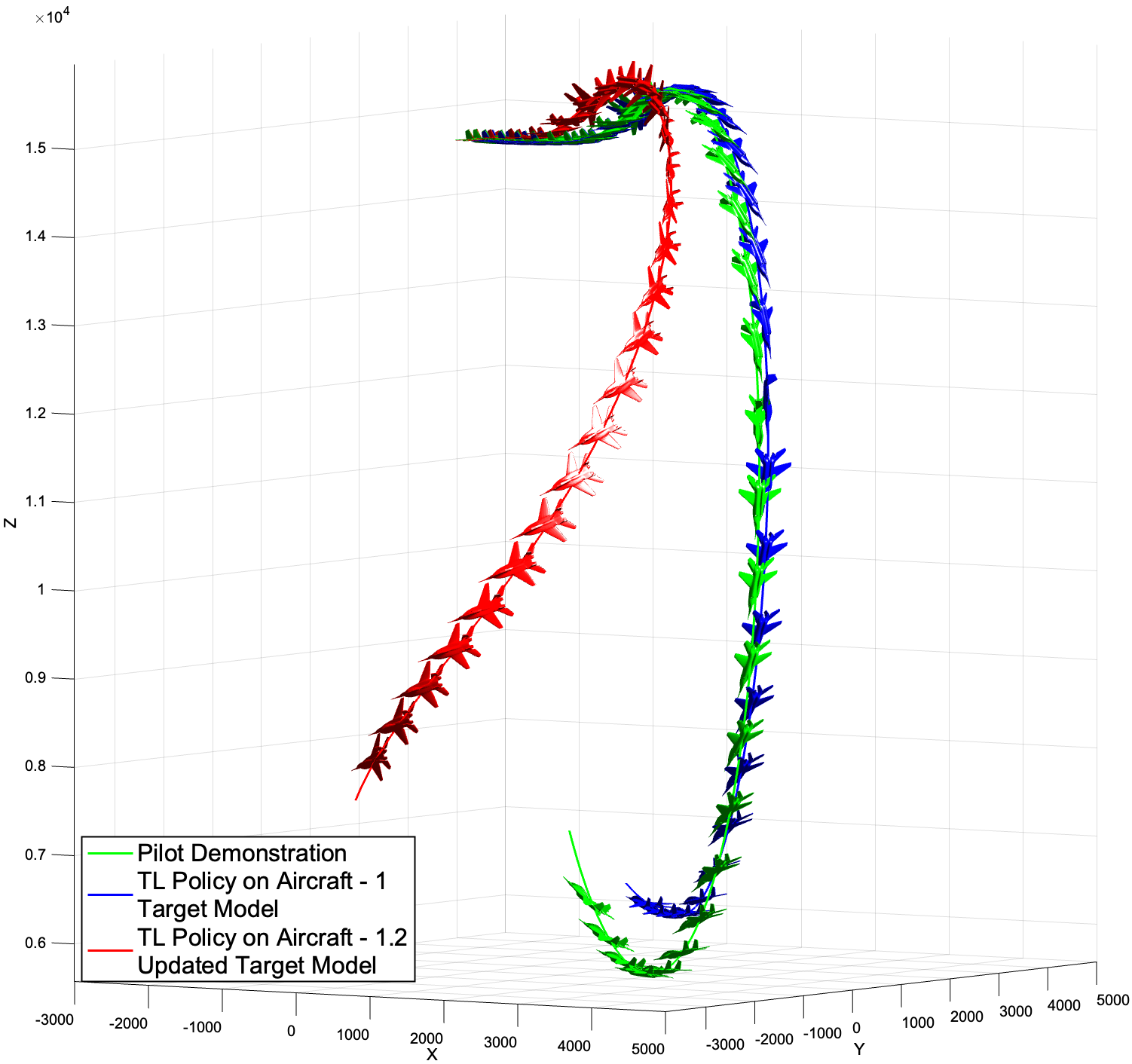}
         \caption{Split-S maneuver}
         \label{fig:splits_tl_heavy_manv}
     \end{subfigure}
     \hfill
    \begin{subfigure}[b]{0.49\textwidth}
         \centering
         \includegraphics[width=\textwidth]{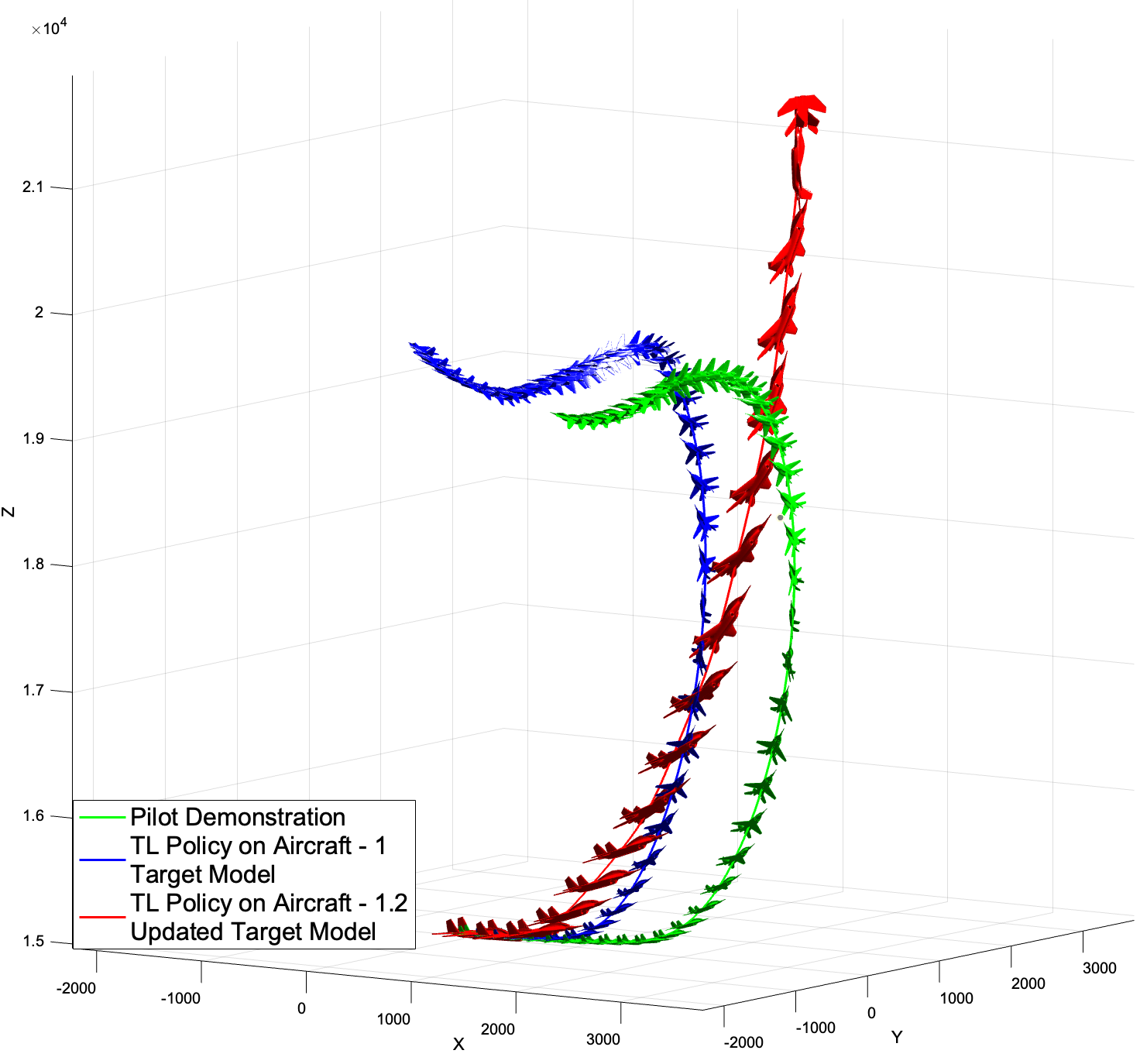}
         \caption{Chandelle maneuver}
         \label{fig:chand_tl_heavy_manv}
     \end{subfigure} 
     \end{adjustbox}
  \caption{Visualization of aircraft trajectories using the Transfer Learning (TL) policy on Aircraft - 2.1 Updated Target Model. The graph indicates the failure of the TL policy to accurately complete the maneuver due to the absence of specific training for this aircraft.}
 \label{fig:tl_heavy_manv}
\end{figure}

The effectiveness of the proposed transfer learning approach was evaluated by increasing the weight of the target aircraft used for transfer learning and observing the resulting performance on updated target aircraft dynamics. The evaluation revealed that the transfer learning policy, which was originally trained on a different aircraft model, was sensitive to the changes in dynamics and limitations of the updated target aircraft. As depicted in Fig. \ref{fig:tl_heavy_manv}, although the initial stages of the maneuvers appeared smooth, it was observed that the aircraft ultimately failed to complete the maneuvers. These results highlight the challenges of transferring a policy across significantly different aircraft dynamics and the need for further research to address these limitations in transfer learning for pilot behavior modeling.

\subsection{Problem Formulation}

\begin{figure*}[t]
     \centering
     \begin{subfigure}[b]{0.49\textwidth}
         \centering
         \includegraphics[width=0.85\textwidth]{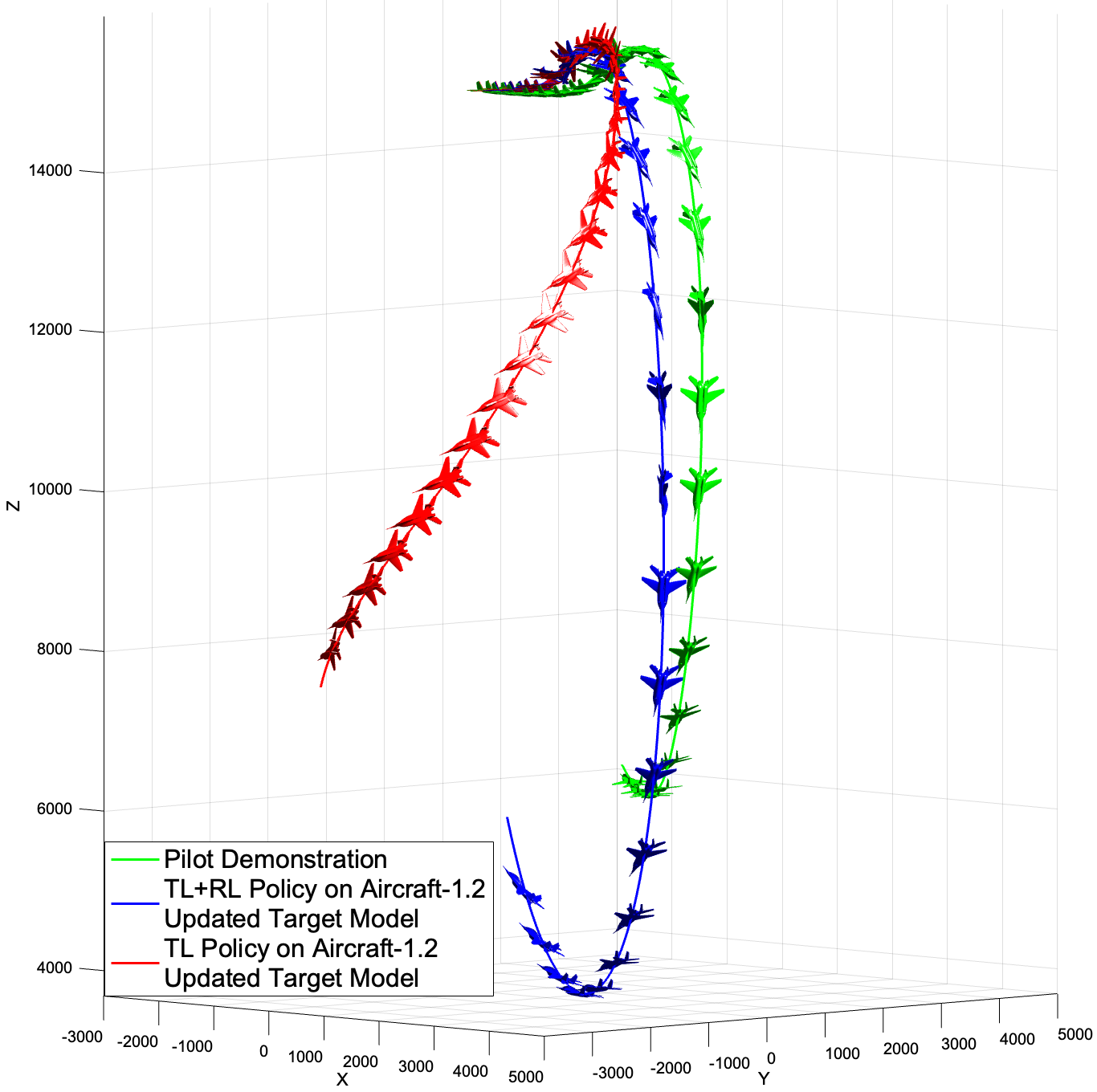}
         \caption{Split-S maneuver}
         \label{fig:splits_rl_heavy_manv}
     \end{subfigure}
     \hfill
    \begin{subfigure}[b]{0.49\textwidth}
         \centering
         \includegraphics[width=0.85\textwidth]{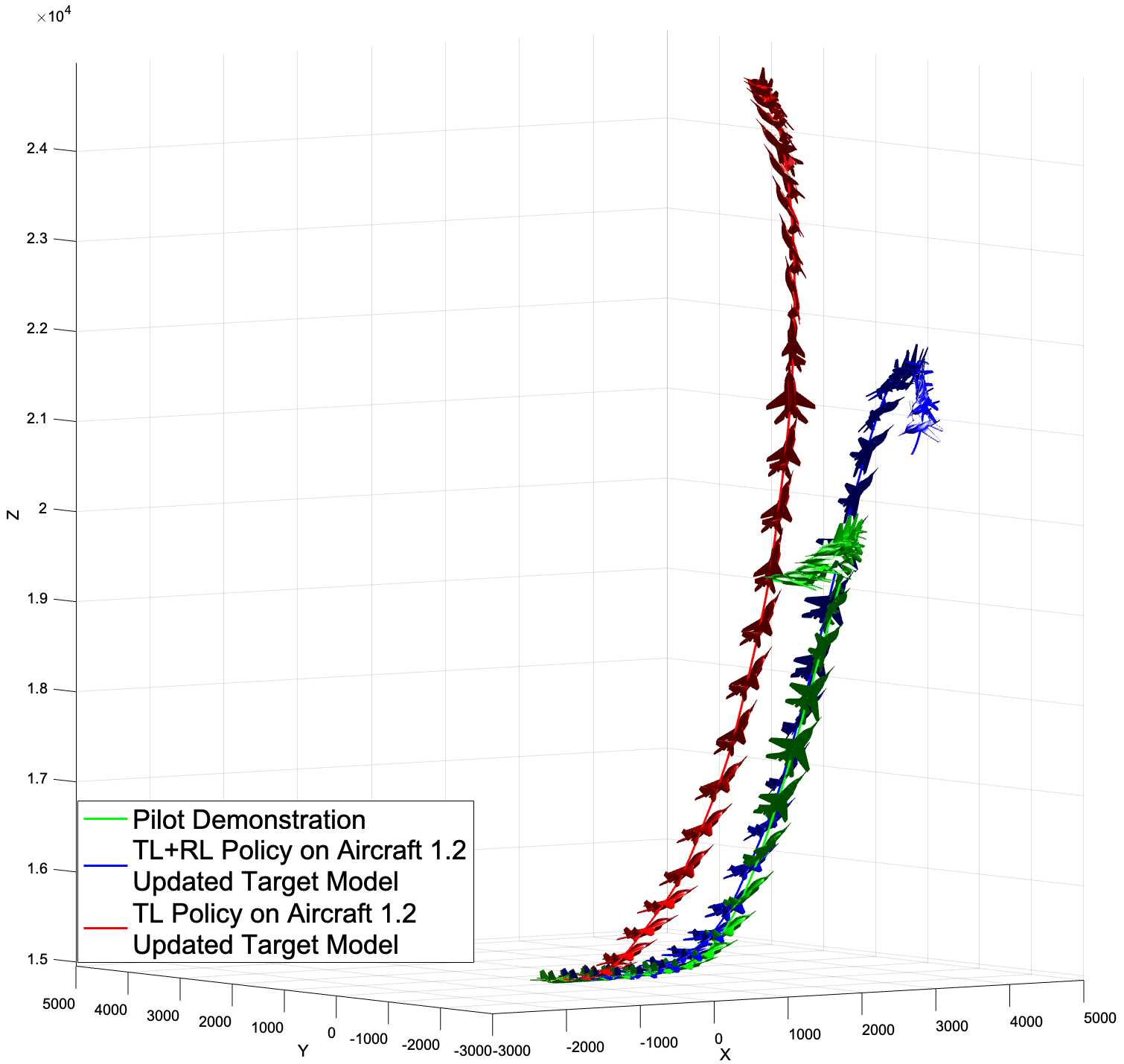}
         \caption{Chandelle maneuver}
         \label{fig:chand_rl_heavy_manv}
     \end{subfigure} 
 \caption{Visualization of aircraft trajectories using different policy approaches. The illustration demonstrates the performance discrepancy between the TL policy, and the combined RL and TL policy. The successful completion of the maneuver with high precision when using the combined RL and TL policy underscores the efficiency and adaptability of this hybrid model.}
\label{fig:rl_heavy_manv}
\end{figure*}

\begin{figure}[H]
    \centering    \includegraphics[width=0.90\columnwidth]{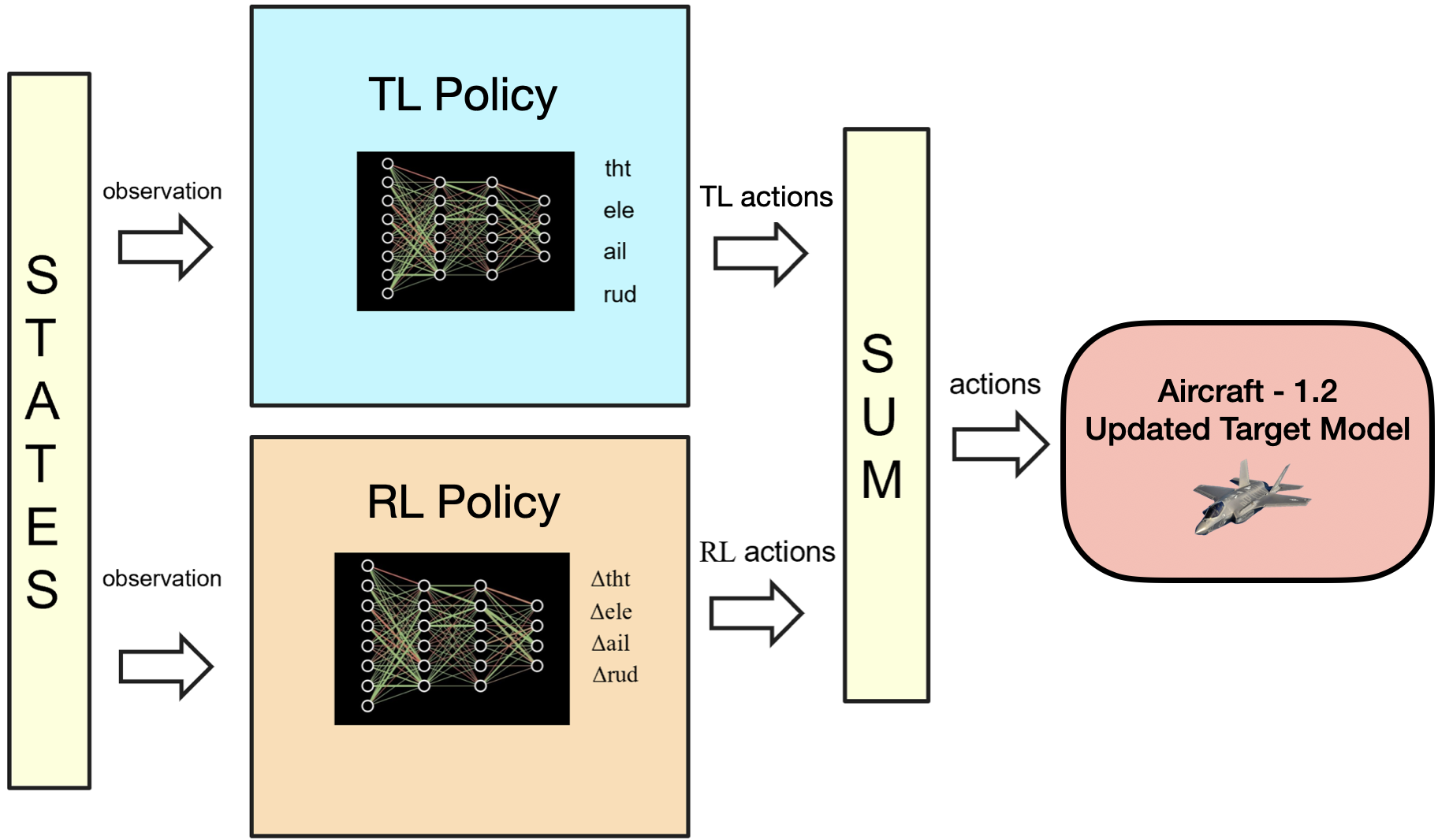}
    \caption{Schematic representation of the interaction between a Reinforcement Learning (RL) model and an Transfer Learning (TL) model for aircraft control. The TL model, with its fixed parameters, provides base control inputs for the aircraft. Concurrently, the adaptable RL model adjusts its parameters to minimize errors in the TL model outputs. The final control inputs are calculated as the sum of outputs from both models.}
    \label{fig:ILsumRL}
\end{figure}

\label{sec:RLformulation}
Reinforcement learning (RL) has been widely applied to a variety of tasks, but its effectiveness in problems that involve highly intricate and sensitive dynamics, such as aircraft control, remains questionable. One major challenge in utilizing RL for aircraft control is the high computational cost of training an agent from scratch, which often results in sub-optimal performance. Instead of training a RL model from scratch to learn the complex dynamics of an aircraft, we propose to fine-tune an existing model obtained through tranfer learning (TL) using the RL approach. The TL model is expected to provide control inputs for the new aircraft, but its parameters will be fixed and remain constant during the fine-tuning process. The RL model, on the other hand, is expected to reduce the errors that occur in the TL model's outputs by adjusting its parameters. The final control inputs for the new aircraft will be the sum of the outputs from the RL model and the TL model, as illustrated in Figure \ref{fig:ILsumRL}. This approach aims to improve the performance of the TL model by leveraging the adaptability of the RL method while maintaining the stability provided by the TL model.

In the RL framework, the problem is formulated as a Markov Decision Process (MDP) \cite{sutton2018reinforcement}, where the goal is to find an optimal policy that maximizes long-term performance. In our MDP formulation, the state space $S$ consists of the aircraft states, and the action space $A$ is continuous, bounded by the aircraft's control input limits, as $ A_{RL} = [\delta_{tht}, \delta_{ele}, \delta_{ail}, \delta_{rud}]$, which includes the control inputs for thrust ($\delta_{tht}$), elevator ($\delta_{ele}$), aileron ($\delta_{ail}$), and rudder ($\delta_{rud}$). The transition dynamics of the agent are governed by the aircraft dynamics. The policy function $\pi(A|S;\theta)$ parameterized by $\theta$ determines the agent's actions.

The control inputs for the aircraft are determined by a combination of the outputs from the TL model, denoted as $A_{TL}$, and the RL model, denoted as $A_{RL}$, as expressed in Eq. \ref{eq:ILsumRL}. To control the influence of the RL model on the final control inputs, a coefficient $C_{RL}$ is introduced. The value of $C_{RL}$ is determined based on the deviation between the dynamics of the aircraft used to train the TL model and the updated target aircraft. If the deviation is significant, a higher value of $C_{RL}$ is chosen to provide more correction to the TL model's output. Conversely, if the deviation is low, a smaller value of $C_{RL}$ is used to minimize the correction applied by the RL model. This dynamic adjustment of the RL model's influence allows the system to adapt to variations in the aircraft's dynamics.

\begin{equation}
\label{eq:ILsumRL}
A_{TL+RL} = A_{TL} + C_{RL}A_{RL} \text{  and  } C_{RL} \in (0, 1]
\end{equation} 
\subsection{Model Selection}

In this study, we employed the Twin Delayed Deep Deterministic Policy Gradient (TD3) algorithm \cite{fujimoto2018addressing} as the off-policy algorithm for reinforcement learning (RL) training. The use of an off-policy algorithm allows for parallel training of the agent, which is essential for efficient computations considering the large number of simulations required in the training process. The simulations were conducted in a Matlab/Simulink environment using a high-fidelity aircraft model, resulting in low run-time for each simulation. Among the available off-policy algorithms, TD3 was selected due to its ability to support parallel training and provide stable learning. This choice ensures both efficient training and accurate model representation. TD3 is an extension of the Deep Deterministic Policy Gradient (DDPG) algorithm that addresses some of its limitations \cite{fujimoto2018addressing}.

\begin{figure}[H]
    \centering
    \includegraphics[width=\columnwidth]{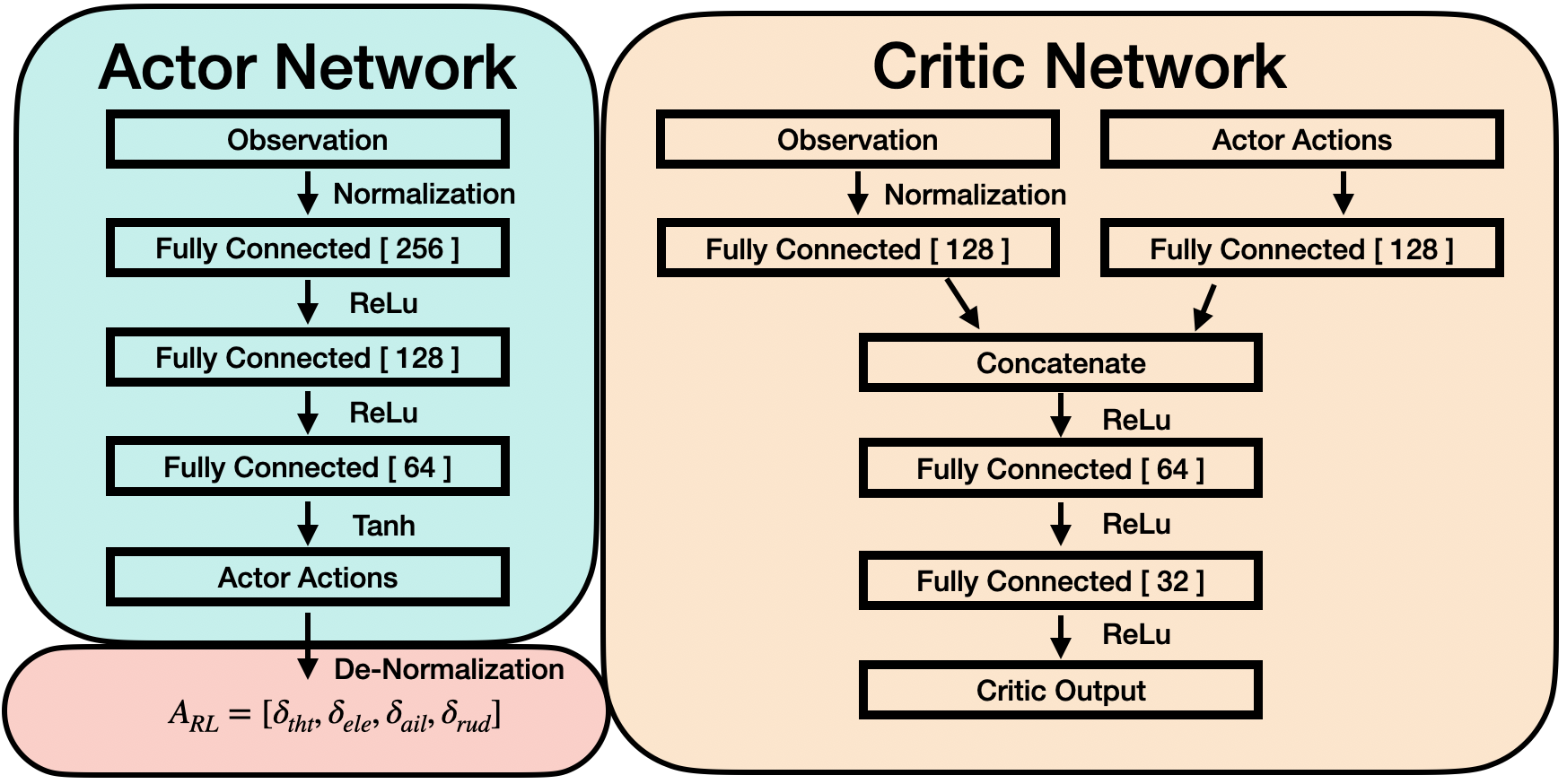}
    \caption{Structure of the Actor-Critic Network used in TD3 algorithm. The Actor network takes normalized observations based on the aircraft's states, passes them through fully-connected layers, and then generates continuous actions within the control input limits using a hyperbolic tangent function. The actions are then de-normalized to match the aircraft stick limits. The Critic network estimates the value function. The numbers within the network represent the number of neurons in each layer.}
    \label{fig:actorcritic}
\end{figure}

Our reinforcement learning (RL) policy consists of two main components: an actor network and a critic network, as depicted in Figure \ref{fig:actorcritic}. To facilitate learning, the observations are first normalized to the (-1,1) interval based on the aircraft's state limits. The normalized observations are then passed through the fully-connected layers of the actor network. The actor network generates continuous actions within the aircraft's control input limits by utilizing a hyperbolic tangent (Tanh) activation function. The Tanh function maps the output to the interval of (-1,1), which corresponds to the range of valid actions for the aircraft. Finally, the actor actions are de-normalized to conform to the aircraft's stick limits, ensuring that the RL output adheres to the defined control input range.

\subsection{Reward Function}
In this study, the performance of the agent is assessed using a reward function, as depicted in Eq. \ref{eq:RLreward}. The reward function is designed to capture the deviation of the agent's maneuvers from a reference pilot data. It is important to note that if the difference between the reference pilot and the maneuver's roll, pitch, and yaw rates ($P, Q, R$) exceeds a significant threshold, the simulation is terminated. This termination condition encourages the agent to stay within acceptable limits and reinforces the importance of closely aligning its maneuvers with the reference pilot data.

\begin{equation}
\label{eq:RLreward}
 R_t =  \Delta t - \Delta t(\Delta P_t+\Delta Q_t+\Delta R_t )
\end{equation}

To incentivize the RL agent to successfully complete the maneuver, a temporal component, $\Delta t$, was incorporated into the reward function. This component ensures that the agent is motivated not only to closely follow the reference pilot data but also to complete the maneuver within a specified time frame. By including this temporal aspect, the agent is encouraged to execute the maneuver efficiently, further maximizing its reward.

\subsection{Training of the Reinforcement Learning Agent and Results}

The RL agent was trained using specific parameters, as outlined in Table \ref{tab:td3table}. The selection of these parameters was based on iterative experimentation to achieve a balance between exploration and exploitation during the learning process. 

\begin{table}
\centering
\resizebox{\columnwidth}{!}{%
\begin{tabular}{|l|l|l|c|}
\hline
$C_{RL}$ & 0.1 & Exploration Model & Ornstein-Uhlenbeck Action Noise \\ \hline
Exploration Mean & 0 & Discount Factor & 0.99 \\ \hline
Exploration Std & 0.1 & Mini Batch Size & 128 \\ \hline
Exploration Std Decay Rate & $10^5$ & Target Smooth Factor & 0.05 \\ \hline
Target Update Frequency & 4 & Policy Update Frequency & 8 \\ \hline
\end{tabular}%
}
\caption{Parameters for TD3 Training}
\label{tab:td3table}
\end{table}

To minimize the initial deviation from the desired trajectory, a small coefficient ($C_{RL}$) was intentionally chosen for the RL agent. This decision allows the RL agent to make subtle adjustments to the control inputs generated by the TL model, rather than introducing drastic changes. By leveraging the prior knowledge encoded in the TL model and using the RL agent as a corrective measure, the overall performance is enhanced, while still adapting to the specific requirements of the task.

\begin{figure}[H]
     \centering
     \begin{adjustbox}{minipage=\linewidth}
     \begin{subfigure}[b]{0.49\textwidth}
         \centering
         \includegraphics[width=\textwidth]{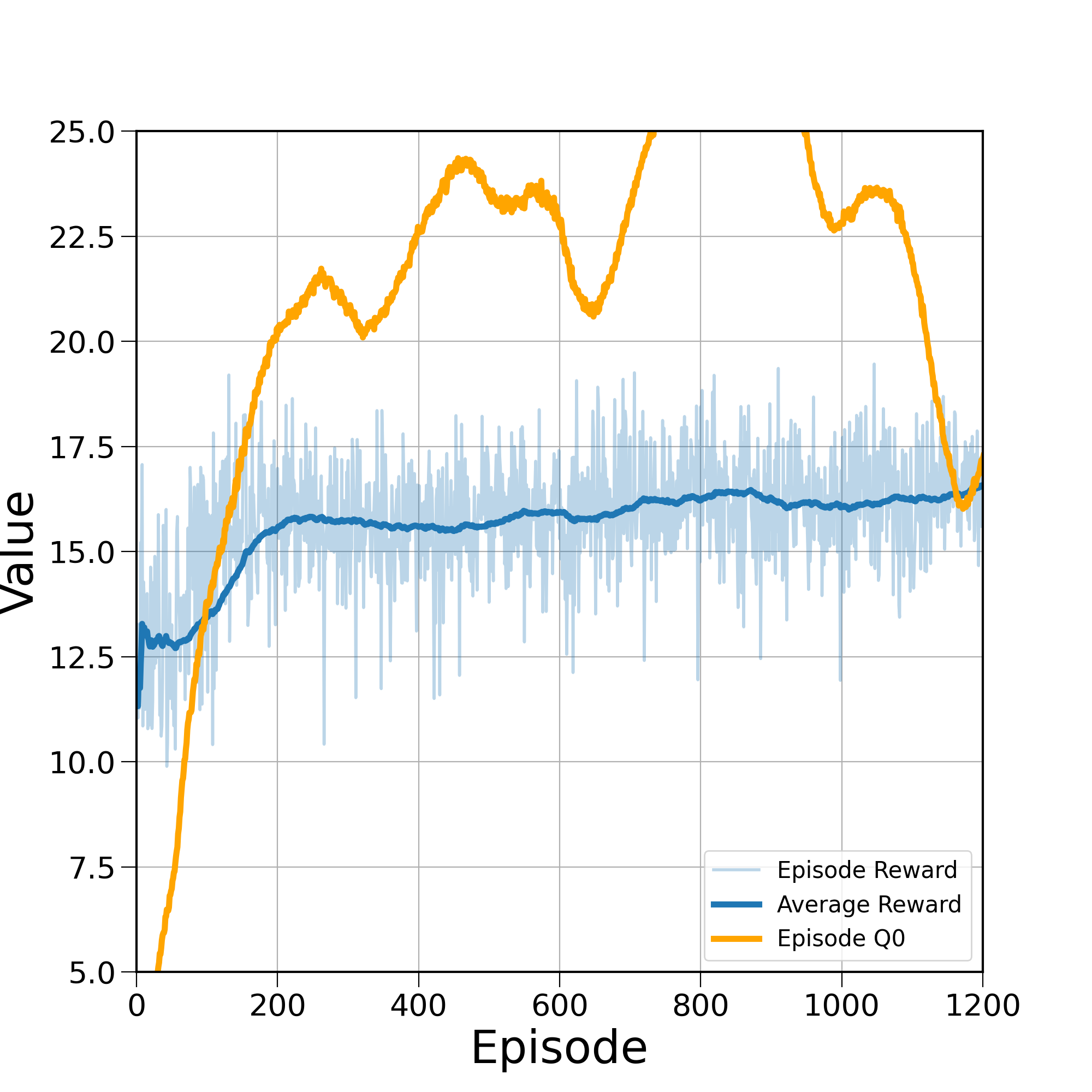}
         \caption{Split-S maneuver}
         \label{fig:splits_rl_hist}
     \end{subfigure}
     \hfill
    \begin{subfigure}[b]{0.49\textwidth}
         \centering
         \includegraphics[width=\textwidth]{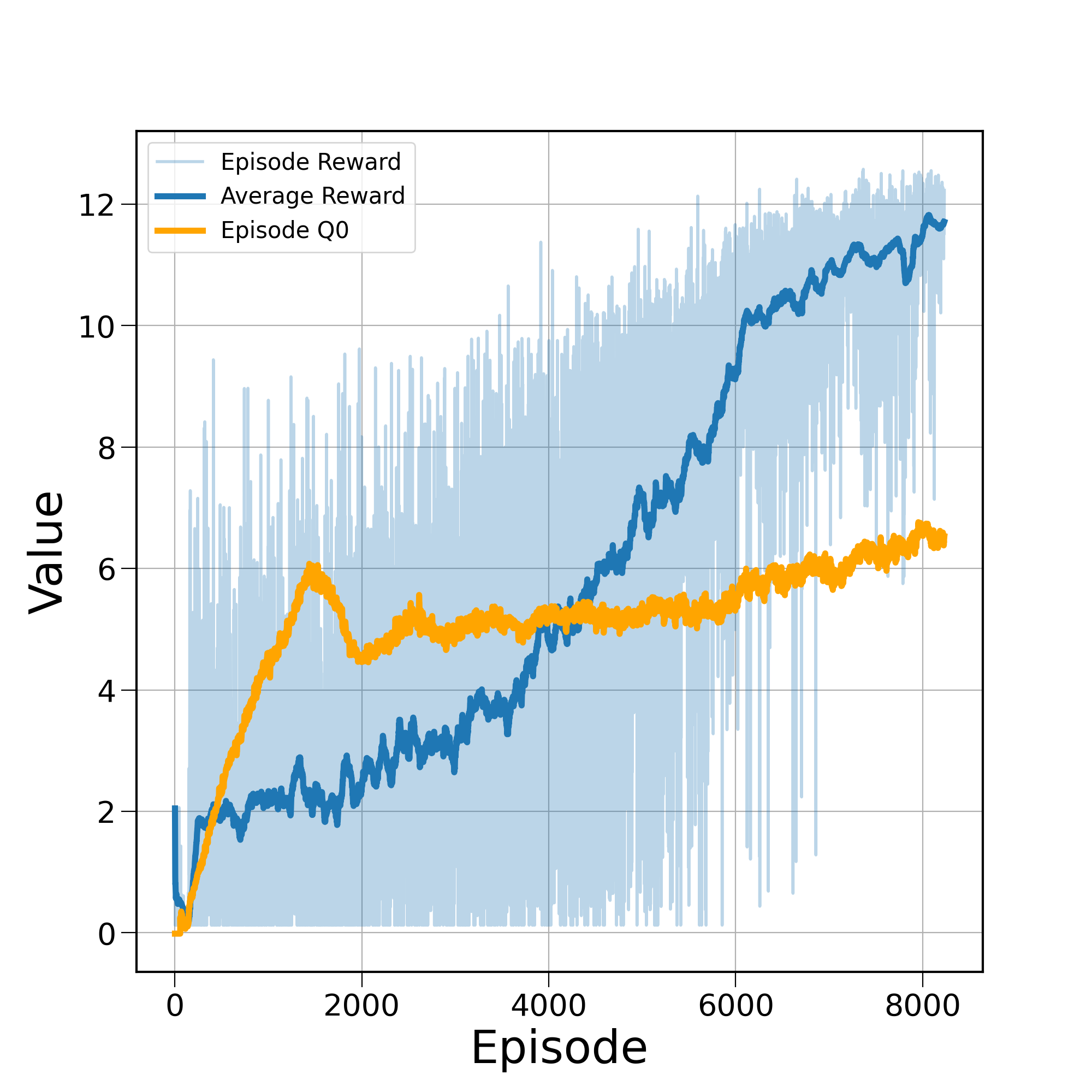}
         \caption{Chandelle maneuver}
         \label{fig:chand_rl_hist}
     \end{subfigure} 
     \end{adjustbox}
 
 \caption{Depiction of the reward history and Q0 values during the TD3 reinforcement learning training }
 \label{fig:RL_training_history}
\end{figure}

The evolution of the reward history and $Q$ values during the TD3 reinforcement learning training is depicted in Figure \ref{fig:RL_training_history}. The RL agent was observed to progressively improve over the training epochs, as reflected in the increasing trend in the reward. It is important to emphasize that the primary goal of this training phase was to replicate the early stages of the maneuvers with high precision. This focus ensures that any subsequent stages of the maneuver are not affected by a snowball effect resulting from an erroneous initial state. This is particularly significant as any deviation from the expected trajectory terminates the simulation, resulting in a low reward score. Further, adding a temporal component to the reward function helped encourage the RL agent to strive for efficient maneuver completion. The reward mechanism ensures that the agent understands that the only way to increase the reward is by not deviating from the reference pilot data.  It is notable that despite the strict reward conditions, both maneuvers could be successfully learned using the proposed methodology.

During the testing phase, the performance of the RL policy was assessed based on its ability to execute the maneuvers accurately and efficiently. The trajectories of the aircraft during the Split-S and Chandelle maneuvers, as depicted in Fig. \ref{fig:rl_heavy_manv}, serve as evidence of the effectiveness of the combined reinforcement learning (RL) and transfer learning (TL) policy. Notably, the hybrid RL and TL policy exhibited a high level of precision and successfully completed the maneuvers, highlighting its adaptability and efficiency compared to the standalone TL policy. 

Additionally, the training strategy presented in our research  remarkably reduce the need for extensive simulations, which can be seen in the results portrayed in the Figure \ref{fig:RL_training_history}. Our approach leverages a TL policy that encapsulates the dynamics of the aircraft, thus enabling swift adaptation to design modifications. As a result, it takes merely 1-2 hours of training to develop an efficient pilot model, marking a significant advancement in simulation efficiency.

\section{Conclusion}

In this paper, we proposed a novel approach for agile maneuver generation in aircraft control, combining behavior cloning, transfer learning, and reinforcement learning techniques. 
By leveraging expert demonstrations, transfer learning, and the integration of reinforcement learning, we developed robust and adaptable agile maneuver models capable of executing complex aerobatic maneuvers with high precision. Our results demonstrated the effectiveness of the proposed methodologies in achieving accurate and efficient maneuver execution, by using only a limited amount of expert pilot demonstration data.

\appendix

\bibliographystyle{elsarticle-num} 
\bibliography{ref}





\end{document}